\documentclass[10pt,twocolumn,letterpaper]{article}

\usepackage[pagenumbers]{cvpr}      % To produce the REVIEW version
\usepackage{algorithm}
\usepackage{algpseudocode}
\usepackage{multirow}            % experiments: tables
\usepackage[export]{adjustbox}   % experiments: valign=c in qualitative figure grids
\usepackage{graphicx}
\usepackage{booktabs}
\usepackage{multirow}
\definecolor{cvprblue}{rgb}{0.21,0.49,0.74}
\usepackage[pagebackref,breaklinks,colorlinks,allcolors=cvprblue]{hyperref}

\def\confName{CVPR}
\def\confYear{2027}

\title{PartPO: Part-Level Policy Optimization for Structured Shape Program Generation}

\author{
  Hannuo Zhang\textsuperscript{1,2},
  Zhixiang Chi\textsuperscript{3},
  Xuanchen Li\textsuperscript{4}, \\
  Yiqun Zhou\textsuperscript{5},
  Yang Wang\textsuperscript{1,2},
  Xinxin Zuo\textsuperscript{1} \\
  \textsuperscript{1}Concordia University \quad
  \textsuperscript{2}Mila -- Quebec AI Institute \quad
  \textsuperscript{3}University of Toronto \\
  \textsuperscript{4}McGill University \quad 
  \textsuperscript{5}Zhejiang University \\
  \texttt{hannuo.zhang@mail.concordia.ca}
}

\begin{document}
\maketitle
%% ============================================================
%% Abstract
%% ============================================================

\begin{abstract}
We present PartPO, a reinforcement learning method for generating structured shape programs from 3D point clouds. Existing shape-to-program models are trained with supervised learning on procedurally generated data and generalize poorly to more diverse shapes, while reinforcement learning from execution feedback assigns a single scalar reward to a long program that builds many parts, making credit assignment difficult. PartPO instead builds on a core property of the task: shape programs are self-describing, with each semantically named part produced by its own code block. It segments each generated program by its parts and applies part-level rewards only to the code of the corresponding part, a category reward only to the object name, and a shape-level reward to the whole program. The required part supervision is obtained without manual annotation, by grounding part names on target meshes with an open-vocabulary segmentation model prompt-tuned on the model's own outputs. Applied to MeshCoder, PartPO improves reconstruction accuracy and execution success on ShapeNetCore and ABO over the base model, rejection-sampling fine-tuning and trajectory-level RL, and generalizes to the unseen HSSD dataset. Code will be released at \url{https://github.com/HannuoZ/PartPO}.
\end{abstract}
%% ============================================================
%% Section 1: Introduction
%% Citation keys are placeholders; keep them consistent with experiments.tex.
%% ============================================================

\section{Introduction}
\label{sec:intro}

\begin{figure}[t]
\centering
\includegraphics[width=\columnwidth]{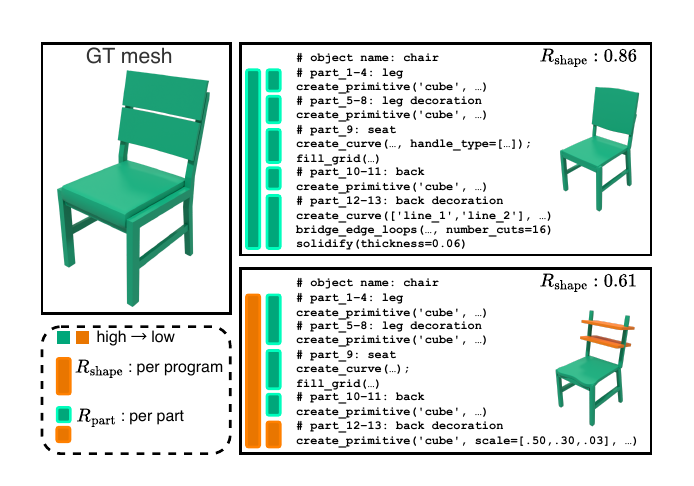}
\caption{Two programs generated for the same chair. A shape-level reward $R_{\text{shape}}$ scores each program as a whole (left bars): the lower program is penalized for its misplaced back decoration, and so are its correct legs, seat and back. PartPO adds a part-level reward $R_{\text{part}}$ (right bars) that scores each part block by its own geometry, so only the faulty block is penalized.}
\label{fig:teaser}
\end{figure}

Generating structured shape programs from 3D perceptual inputs aims to translate a point cloud or mesh into executable code that reconstructs the observed object~\cite{dai2025meshcoder,kolodiazhnyi2025cadrille}. Unlike meshes or implicit fields, a shape program is compact, editable and interpretable, which makes it a valuable representation for shape editing, reverse engineering and 3D structure understanding. Recent work frames this task, like related CAD generation problems~\cite{li2025cadllama,guan2025cadcoder}, as code generation with multimodal large language models (MLLMs). MeshCoder~\cite{dai2025meshcoder}, for example, defines an expressive domain-specific language (DSL) in Blender Python whose programs are self-describing: they name the object and each of its parts, and pair every part name with the code block that produces that part's geometry. Such models are trained with supervised learning on 3D object--program pairs that are generated procedurally~\cite{raistrick2024infinigen}.

However, these methods struggle to generalize to more diverse or out-of-distribution data, such as human-authored 3D assets. Paired training data are limited because they must be generated procedurally, and the model is supervised only by code: it never observes the mesh obtained by executing its programs during training. Token-level losses encourage programs that resemble the training code, but they do not tell the model whether the executed shape actually matches the input.

To improve the visual program inference capability of MLLMs, recent work uses reinforcement learning (RL), typically with GRPO~\cite{shao2024deepseekmath,guo2025deepseekr1,yu2025dapo}, and rewards each program by comparing its executed output with the target image or shape~\cite{rodriguez2025rlrf,kolodiazhnyi2025cadrille,guan2025cadcoder}. These methods, however, face a credit assignment challenge: for long, structured and compositional programs, they assign a single scalar reward to each trajectory. Because this reward is sparse and delayed, available only once the whole program has been generated and executed, it is hard to identify which tokens contributed positively or negatively~\cite{guo2025spo,cui2025prime,kazemnejad2025vineppo}. A program with accurate legs and a misplaced backrest receives one score for all of its tokens (Fig.~\ref{fig:teaser}), so the advantage of each part is confounded by the quality of the others and partial progress goes unrewarded, which prior work identifies as a source of imprecise credit assignment and reduced sample efficiency.

To address this challenge, we propose PartPO, an RL method that uses the self-describing structure of shape programs for fine-grained credit assignment. PartPO segments each rollout into its object-name line and named part blocks, and routes each reward only to the tokens that produce what it evaluates: a \textit{part-level reward}, comparing each generated part with the same-named part of the target, acts on the code of that part block; a \textit{category reward} acts on the object-name line; and a \textit{shape-level reward}, measuring geometric and semantic similarity to the input, acts on all tokens to preserve overall reconstruction quality. Because the segments come from the program itself and all rewards are verifiable geometric comparisons, PartPO needs neither a learned reward model nor additional rollouts. Its part supervision requires no manual annotation either: PartPO prompts an open-vocabulary segmentation model~\cite{carion2025sam3} with the part names in MeshCoder's own programs and lifts the resulting multi-view masks to 3D, so it applies to datasets that provide only meshes with category labels.

Applied to MeshCoder, PartPO improves over the base model and outperforms rejection-sampling fine-tuning and trajectory-level RL on ShapeNetCore and ABO. It reduces the Chamfer distance of MeshCoder by 53\% on ShapeNetCore and 60\% on ABO, and raises the execution success rate on ABO from 58.6\% to 97.8\%. Trained only on ShapeNetCore, it also generalizes zero-shot to the unseen HSSD dataset, raising the success rate from 81.3\% to 89.2\% and reducing the Chamfer distance by 14\%.

Our contributions are as follows:
\begin{enumerate}
    \item We propose PartPO, an RL method for structured shape program generation that introduces part-level advantage estimation to overcome the limitations of trajectory-level RL methods.
    \item We design a set of complementary rewards that mirror the program structure, including shape-level, part-level and category rewards, and build part supervision for the part-level reward without manual annotation.
    \item Experiments on three datasets show that PartPO outperforms trajectory- and segment-level RL baselines, and part-level evaluation and ablations confirm that its gains come from more accurate parts and structure-aligned credit assignment.
\end{enumerate}

%% ============================================================
%% Section 2: Related Work
%% Citation keys are placeholders; keep them consistent with introduction.tex and experiments.tex.
%% ============================================================

\section{Related Work}
\label{sec:related}

\paragraph{Structured Shape Representations.}
Primitive-based methods abstract shapes into compact parts: SuperDec~\cite{fedele2025superdec} fits superquadrics to point clouds, PrimitiveAnything~\cite{ye2025primitiveanything} generates primitive assemblies learned from human-crafted abstractions, and PartCrafter~\cite{lin2025partcrafter} generates an object as semantically meaningful part meshes. They output part geometry rather than executable programs. CAD methods instead recover construction sequences: DeepCAD~\cite{wu2021deepcad} models sketch-and-extrude sequences, and CAD-Recode~\cite{rukhovich2025cadrecode}, cadrille~\cite{kolodiazhnyi2025cadrille} and CAD-Llama~\cite{li2025cadllama} generate executable CAD code with language models; for text-to-CAD, CAD-Coder~\cite{guan2025cadcoder} applies GRPO with a Chamfer-distance reward and CADFusion~\cite{wang2025cadfusion} uses preference optimization on rendered models. These operation sequences, however, neither name parts nor model their relations. Shape programs make this structure explicit: Shape2Prog~\cite{tian2019shape2prog} captures repetition and symmetry with loops over primitives, aDSL~\cite{wang2026adsl} designs a DSL with semantic part classes for LLM agents working from text or images, and MeshCoder~\cite{dai2025meshcoder} translates point clouds into Blender Python programs that pair every named part with the code block producing it. Such models are trained mainly by supervised learning on procedurally generated pairs; PartPO adapts them to more diverse data with RL and exploits their part structure for credit assignment.

\paragraph{RL from Execution Feedback.}
RL~\cite{schulman2017ppo,shao2024deepseekmath,guo2025deepseekr1,yu2025dapo} has become a standard way to post-train code models with execution feedback, \eg, with a unit-test critic in CodeRL~\cite{le2022coderl} or multi-turn refinement in RLEF~\cite{gehring2025rlef}. For visual code, the executed output can be compared directly with the target: RLRF~\cite{rodriguez2025rlrf} rewards rendered SVG code by reconstruction and semantic similarity, and cadrille~\cite{kolodiazhnyi2025cadrille} rewards CAD code by its geometric agreement with the target shape. These methods assign one scalar reward per trajectory, which makes it hard to tell which parts of a long, multi-part program are right or wrong.

\paragraph{Fine-Grained Credit Assignment.}
Denser signals for RL of language models usually come at an extra cost. Process reward models score intermediate steps but require human step labels~\cite{lightman2024verify}, labels from Monte Carlo (MC) completion~\cite{wang2024mathshepherd}, or an implicit reward model learned online~\cite{cui2025prime}, and Fine-Grained RLHF~\cite{wu2023finegrained} relies on segment-level human annotations. SPO~\cite{guo2025spo}, VinePPO~\cite{kazemnejad2025vineppo} and TreeRL~\cite{hou2025treerl} estimate segment- or step-level values from additional MC or tree-search rollouts, which in our task means many extra Blender executions; GiGPO~\cite{feng2025gigpo} avoids extra rollouts but assumes that environment states recur across trajectories, as in multi-turn agent tasks. In 3D, MetaSpatial~\cite{pan2025metaspatial} modulates object-level advantages with physics-aware penalties for scene layout generation. PartPO instead takes its segments from the program's own part structure and verifies each part by geometric comparison with the target, requiring no human annotation, learned reward model or extra rollouts. Whereas recent open-world methods segment parts directly in 3D~\cite{liu2025partfield,ma2025find3d}, we follow PartSLIP~\cite{liu2023partslip} in lifting prompt-tuned multi-view 2D predictions to 3D, but tune the prompts on the policy's own executed programs instead of annotated shapes.
%% ============================================================
%% Section 3: Method
%% Requires in the preamble: \usepackage{amsmath,amssymb}
%% Citation keys are placeholders; keep them consistent with the other sections.
%% ============================================================

\section{Method}
\label{sec:method}

\subsection{Overview}
\label{sec:overview}

\begin{figure*}[t]
\centering
\includegraphics[width=\textwidth]{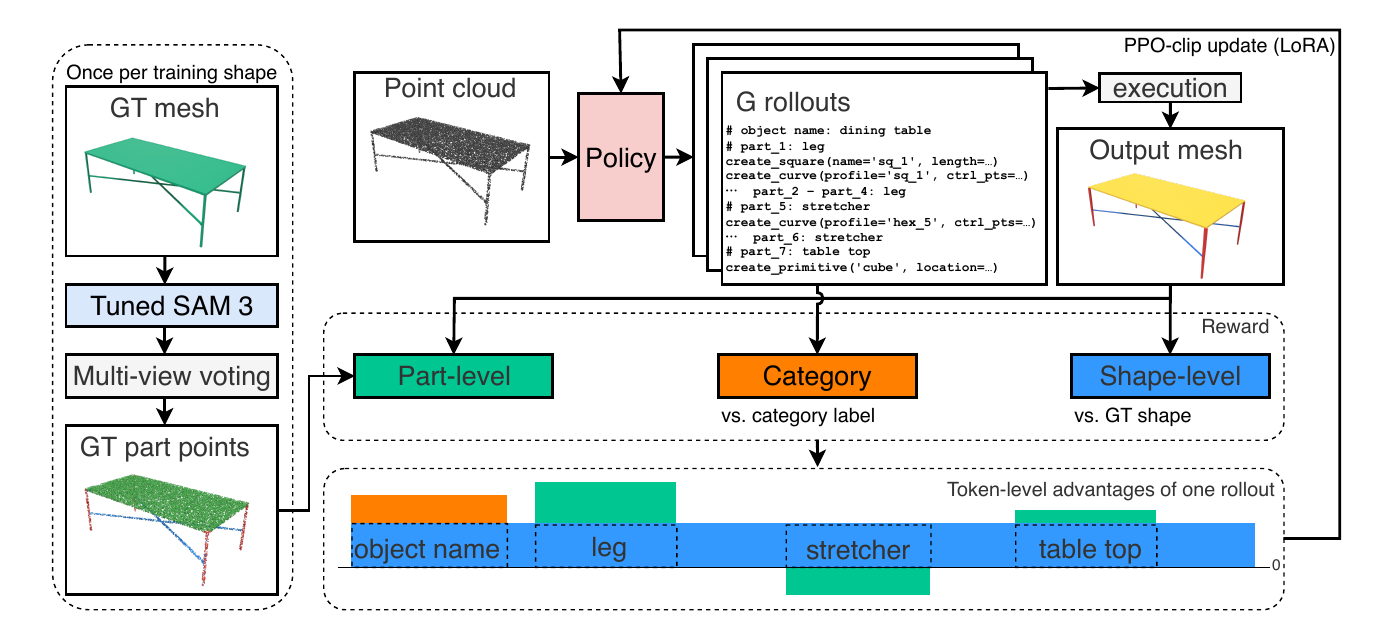}
\caption{Overview of PartPO. For each point cloud, the policy samples G programs, which are executed in Blender and scored by a shape-level reward (vs. GT shape), a category reward (vs. category label) and part-level rewards (vs. GT part points, obtained once per shape with prompt-tuned SAM 3 and multi-view voting, left). Each advantage is added only to the tokens it evaluates (bottom), so the misplaced stretcher is penalized without penalizing the other correct parts.}
\label{fig:pipeline}
\end{figure*}

\paragraph{Shape-to-Code Generation.}
Given a point cloud $\mathcal P\in\mathbb R^{N\times6}$ (coordinates and normals), the goal is to generate an executable program $y=(y_1,\dots,y_T)$ whose execution yields a mesh $\mathcal M(y)$ that reconstructs the input. MeshCoder~\cite{dai2025meshcoder} addresses this task with a language model that writes Blender Python code. A point cloud encoder $E$ maps $\mathcal P$ to $S$ shape tokens $Z=E(\mathcal P)$, which are inserted between a text prefix and suffix, $x=[\text{prefix};Z;\text{suffix}]$, and a Llama-3.2-1B model~\cite{grattafiori2024llama3} generates $y$ by next-token prediction. The output program has a fixed structure: a preamble, an object-name line \texttt{\# object name: <name>}, and a sequence of part blocks. The $k$-th part block $b_k$ starts with a header \texttt{\# part\_k: <part name>} and extends to the next header; we denote its part name by $n(b_k)$ and the sub-mesh it produces by $\mathcal M_k$. Blocks with the same name form one part, $\mathcal M_n=\bigcup_{k:n(b_k)=n}\mathcal M_k$. MeshCoder is trained by supervised fine-tuning on synthetic shape--program pairs from Infinigen-Indoor~\cite{raistrick2024infinigen}, and its reconstructions degrade on human-authored assets such as ShapeNetCore~\cite{chang2015shapenet}, ABO~\cite{collins2022abo} and HSSD~\cite{khanna2024hssd}, whose geometry and part layouts differ from the procedural training distribution.

\paragraph{Addressing Credit Assignment.}
Since Blender execution is not differentiable but each training shape has a ground-truth (GT) mesh, we adapt MeshCoder with group relative policy optimization (GRPO)~\cite{shao2024deepseekmath}. Standard GRPO gives one scalar score to all tokens of a rollout, so in a program of over a thousand tokens a well-built part and a misplaced one receive identical credit. PartPO instead parses each rollout into its object-name line and part blocks, compares each executed part with the same-named GT part (Sec.~\ref{sec:grounding}), and adds each score only to the advantages of the tokens that produced what it evaluates.

\paragraph{PartPO.}
An overview of PartPO is shown in Fig.~\ref{fig:pipeline}. We use three kinds of reward (Sec.~\ref{sec:rewards}). The shape-level reward $R_{\text{shape}}$ is a scalar that measures how closely the whole generated shape matches the target, both geometrically and semantically. The part-level reward $R_{\text{part}}$ acts on each part block and measures how close the generated part is to the corresponding GT part. The category reward $R_{\text{cat}}$ acts on the object-name line and measures whether the object category is identified correctly. Writing a rollout as $y=(y_{\text{name}},b_1,\dots,b_K)$, where $b_k$ here denotes the $k$-th block and $y_{<k}$ all tokens preceding it, we write the expected part-level reward of block $k$ given its preceding context as $\bar R_k(\theta;x,y)=\mathbb{E}_{\tilde b_k\sim\pi_\theta(\cdot\mid x,\,y_{<k})}R_{\text{part}}(x,y;\tilde b_k)$, and define the expected shape-level, category and part-level rewards of an input $x$ as
\begin{equation*}
\begin{aligned}
\mathcal{J}_{\text{shape}}(\theta;x)&=\mathbb{E}_{y\sim\pi_\theta(\cdot\mid x)}R_{\text{shape}}(x,y),\\
\mathcal{J}_{\text{cat}}(\theta;x)&=\mathbb{E}_{y_{\text{name}}\sim\pi_\theta(\cdot\mid x)}R_{\text{cat}}(x,y_{\text{name}}),\\
\mathcal{J}_{\text{part}}(\theta;x)&=\mathbb{E}_{y\sim\pi_\theta(\cdot\mid x)}\textstyle\sum_{k=1}^{K(y)}\bar R_k(\theta;x,y).
\end{aligned}
\end{equation*}
The training objective is
\begin{equation}
\mathcal{J}_{\text{PartPO}}(\theta)=\mathbb{E}_{x\sim\mathcal D}\big[\mathcal{J}_{\text{shape}}+\lambda_c\,\mathcal{J}_{\text{cat}}+\mathcal{J}_{\text{part}}\big](\theta;x),
\label{eq:objective}
\end{equation}
where
\begin{equation}
\begin{split}
R_{\text{part}}(x,y;b_k)={}&\lambda_1R_{\text{sem}}\big(x,y;n(b_k)\big)\\
&+\lambda_2R_{\text{geo}}(x,b_k).
\end{split}
\label{eq:part_reward}
\end{equation}

We optimize Eq.~\eqref{eq:objective} with GRPO. For each input $x$, we sample a group of $G$ programs $\{y^{(i)}\}_{i=1}^G$ from $\pi_{\theta_{\text{old}}}$ and normalize the shape-level reward $R^{(i)}_{\text{shape}}=R_{\text{shape}}(x,y^{(i)})$ within the group,
\begin{equation}
A_o^{(i)}=\frac{R^{(i)}_{\text{shape}}-\operatorname{mean}_jR^{(j)}_{\text{shape}}}{\operatorname{std}_jR^{(j)}_{\text{shape}}+\epsilon}.
\label{eq:shape_adv}
\end{equation}
The part-semantic, part-geometric and category rewards are converted into advantages $A_1^{(i)}[n]$, $A_2^{(i)}[k]$ and $A_c^{(i)}$ in the same group-relative manner (Sec.~\ref{sec:rewards}). PartPO combines them into a token-level advantage for the $t$-th token of rollout $i$:
\begin{equation}
\begin{split}
A^{(i)}_t={}&A^{(i)}_o+\mathbf{1}\big[t\in\text{body}(b_{k(t)})\big]\\
&\cdot\Big(\lambda_1A^{(i)}_1\big[n(b_{k(t)})\big]+\lambda_2A^{(i)}_2\big[k(t)\big]\Big)\\
&+\mathbf{1}\big[t\in\text{name}\big]\,\lambda_cA^{(i)}_c,
\end{split}
\label{eq:token_adv}
\end{equation}
where $k(t)$ is the part block containing token $t$. Part headers, the preamble and the end-of-sequence token receive only $A_o$, as do rollouts that fail to execute. With the probability ratio $r^{(i)}_t=\pi_\theta(y^{(i)}_t\mid x,y^{(i)}_{<t})/\pi_{\theta_{\text{old}}}(y^{(i)}_t\mid x,y^{(i)}_{<t})$ and the clipped per-token term
\begin{equation*}
\ell^{(i)}_t(\theta)=\min\Big(r^{(i)}_tA^{(i)}_t,\ \operatorname{clip}\big(r^{(i)}_t,1-\varepsilon,1+\varepsilon\big)A^{(i)}_t\Big),
\end{equation*}
the final objective is the PPO-clip surrogate of GRPO with token-level advantages, normalized by the total number of tokens $N=\sum_{i=1}^{G}|y^{(i)}|$ in the group:
\begin{equation}
\mathcal{J}_{\text{GRPO}}(\theta)=\mathbb{E}\Bigg[\frac1N\sum_{i=1}^{G}\sum_{t=1}^{|y^{(i)}|}\ell^{(i)}_t(\theta)\Bigg],
\label{eq:grpo}
\end{equation}
where the expectation is over $x\sim\mathcal D$ and $\{y^{(i)}\}_{i=1}^G\sim\pi_{\theta_{\text{old}}}(\cdot\mid x)$.
We omit the KL penalty. During RL, the SFT weights and the point cloud encoder are frozen and only newly added LoRA parameters~\cite{hu2022lora} are trained.

\subsection{Rewards for Shape-to-Code Generation}
\label{sec:rewards}

Our rewards are computed automatically by executing the generated program and comparing the result with the GT shape, so no human annotation is needed. They evaluate the program at three granularities: the whole shape, each part, and the object category. Each training shape provides a normalized GT point cloud $\mathcal X$, a GT mesh $\mathcal M^*$, GT part points $\{\mathcal X^*_n\}$ indexed by part name (Sec.~\ref{sec:grounding}), and a category label $\ell$.

\paragraph{Shape-Level Reward.}
We measure geometric fidelity with a truncated Chamfer similarity (TCS). With the truncated kernel $\psi_c(d)=\max(0,1-d/c)$ and $c=\kappa\cdot\operatorname{diag}(\mathcal X)$, where $\operatorname{diag}(\cdot)$ is the bounding-box diagonal, the similarity between point sets $A$ and $B$ is
\begin{equation}
\begin{split}
S_c(A,B)=\frac12\Big[&\frac1{|A|}\sum_{a\in A}\psi_c\big(d(a,B)\big)\\
&+\frac1{|B|}\sum_{b\in B}\psi_c\big(d(b,A)\big)\Big],
\end{split}
\label{eq:tcs}
\end{equation}
where $d(\cdot,\cdot)$ is the nearest-neighbor distance. Truncation bounds the influence of outliers and keeps the reward in $[0,1]$. A semantic term complements it with the cosine similarity of Uni3D~\cite{zhou2024uni3d} embeddings $f(\cdot)$ of the two point clouds, both colored in a uniform gray so that only geometry is compared. A program that raises an error, times out, or produces an empty or degenerate mesh fails a validity gate and receives $r_{\text{fail}}=0$. With $\hat{\mathcal Y}$ sampled from the generated mesh,
\begin{equation}
R_{\text{shape}}=\begin{cases}w_g\,S_c(\hat{\mathcal Y},\mathcal X)\\
\quad+\,w_s\cos\big(f(\hat{\mathcal Y}),f(\mathcal X)\big), & \text{valid},\\ r_{\text{fail}}, & \text{otherwise},\end{cases}
\label{eq:shape_reward}
\end{equation}
where ``valid'' means that $y$ passes the gate.

\paragraph{Part-Level Reward.}
Part-level rewards are computed for valid rollouts and use one-sided truncated similarities with radius $c_p=\kappa_p\cdot\operatorname{diag}(\mathcal X)$. The \emph{part-semantic} reward measures whether a claimed part covers the true one. For each part name $n$, it compares the GT part points $\mathcal X^*_n$ (Sec.~\ref{sec:grounding}) with the generated part $\mathcal M_n$, keeping only the closest $\rho$ fraction of points to tolerate segmentation noise:
\begin{equation}
R_{\text{sem}}(x,y;n)=\frac1{\lceil\rho|\mathcal X^*_n|\rceil}\sum_{p\in\operatorname{top}_\rho(\mathcal X^*_n)}\psi_{c_p}\big(d(p,\mathcal M_n)\big).
\label{eq:part_sem}
\end{equation}
The \emph{part-geometric} reward measures whether a block's geometry lies on the object. It is independent of part names and penalizes extra or misplaced geometry. With $\hat{\mathcal Y}_k$ sampled from the sub-mesh $\mathcal M_k$,
\begin{equation}
R_{\text{geo}}(x,b_k)=\frac1{|\hat{\mathcal Y}_k|}\sum_{p\in\hat{\mathcal Y}_k}\psi_{c_p}\big(d(p,\mathcal M^*)\big).
\label{eq:part_geo}
\end{equation}
The two terms act as recall and precision at the part level. Each is centered within the group, against units of the same part name when that name appears in at least two rollouts and against all units otherwise. It is then divided by the standard deviation over the whole optimization step rather than within the group, which avoids amplifying noise in small groups. This yields $A_1^{(i)}[n]$ and $A_2^{(i)}[k]$; units whose GT part is missing receive zero advantage.

\paragraph{Category Reward.}
The part-level reward is effective only when the model writes part names of the correct category. If the model mistakes the input for another category, its part names do not exist in the GT part vocabulary, and the part-semantic reward becomes uninformative for the entire rollout. We therefore reward the category decision directly. We parse the noun phrase on the object-name line and map its last word to a category $\hat c(y)$; if the mapping fails, $\hat c(y)=\textsc{Unknown}$. Comparing with the label $\ell$,
\begin{equation}
R_{\text{cat}}=\begin{cases}+1, & \hat c=\ell,\\ 0, & \hat c=\textsc{Unknown},\\ -1, & \text{otherwise}.\end{cases}
\label{eq:cat_reward}
\end{equation}
$A_c^{(i)}$ is obtained by subtracting the group mean and dividing by the step-level standard deviation. It is therefore nonzero only in groups that contain both correct and incorrect predictions, where it shifts probability toward the rollouts that name the category correctly.
\begin{figure}[t]
\centering
\setlength{\tabcolsep}{0pt}
\renewcommand{\arraystretch}{0}
\newcommand{\gimg}[1]{\includegraphics[width=0.22\columnwidth]{figures/qualitative/#1}}
\begin{tabular}{cccc}
{\scriptsize View 1} & {\scriptsize View 2} & {\scriptsize View 3} & {\scriptsize 3D part points} \\[4pt]
\gimg{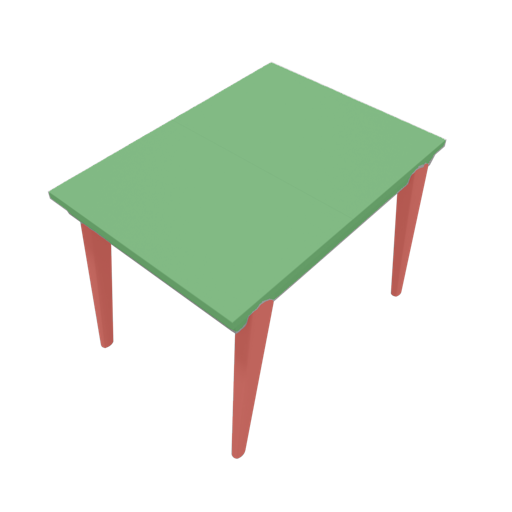} &
\gimg{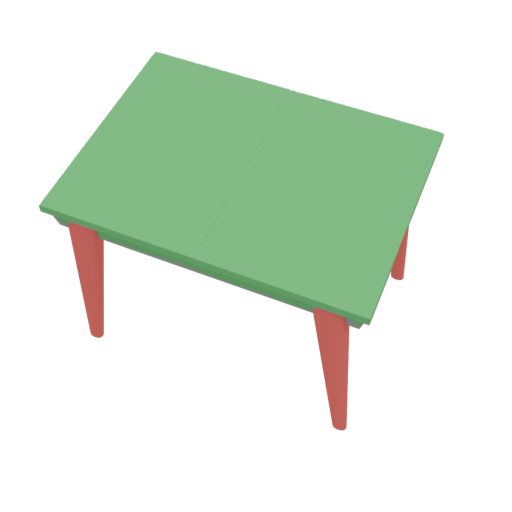} &
\gimg{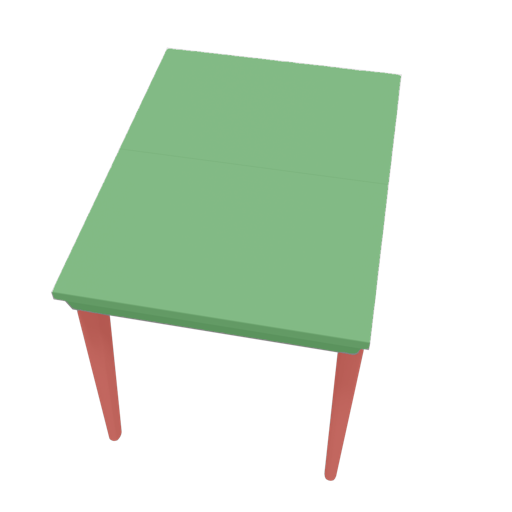} &
\gimg{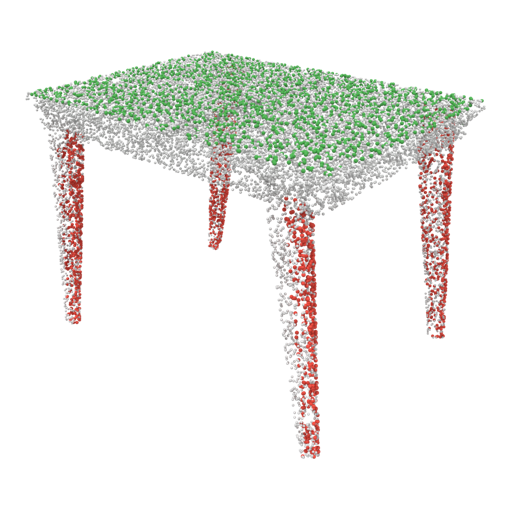} \\
\gimg{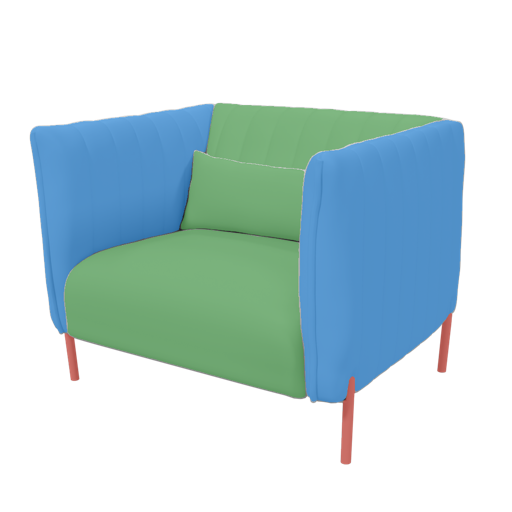} &
\gimg{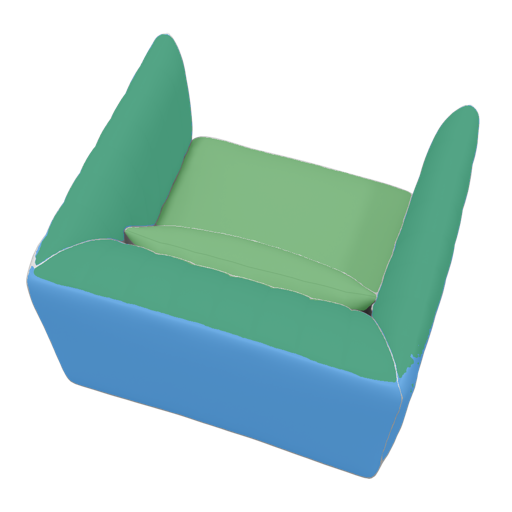} &
\gimg{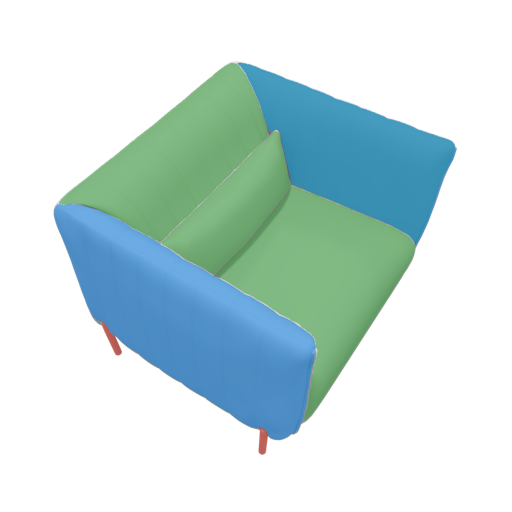} &
\gimg{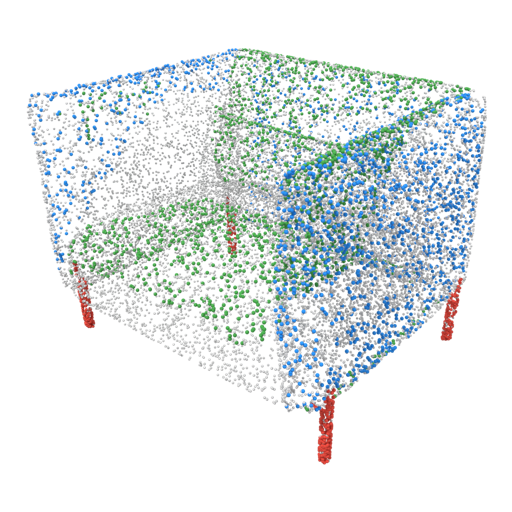} \\
\gimg{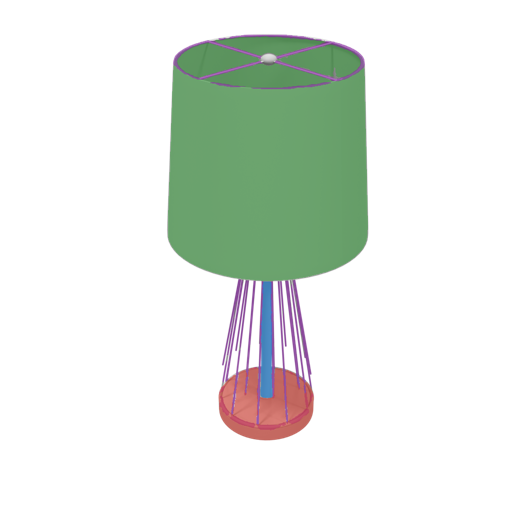} &
\gimg{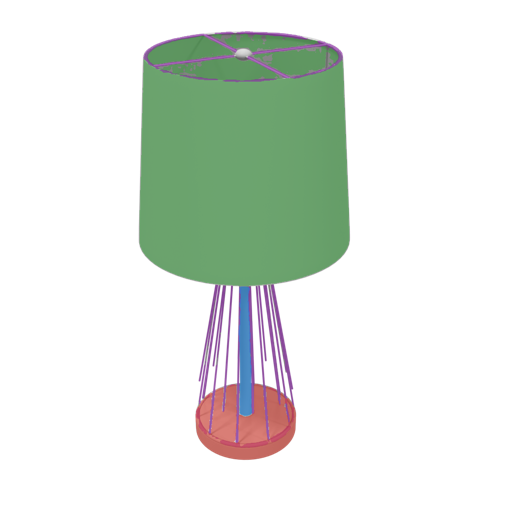} &
\gimg{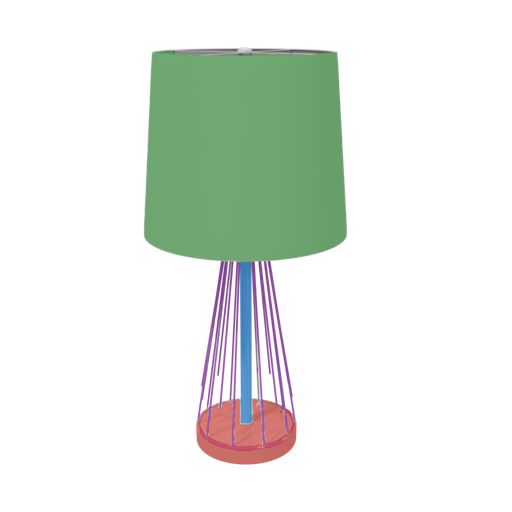} &
\gimg{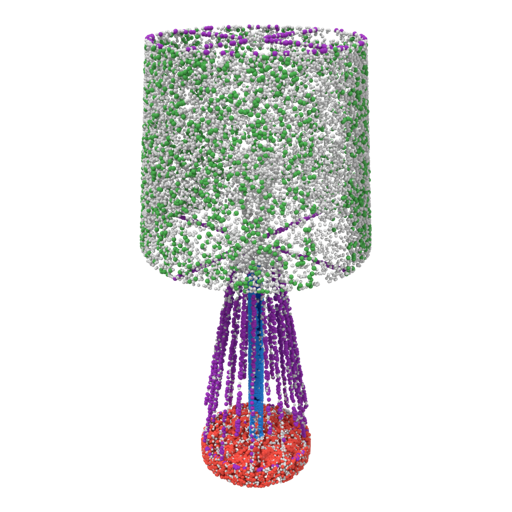} \\
\end{tabular}
\caption{Grounding part names on GT meshes. Columns 1--3: masks predicted by the prompt-tuned SAM~3 on three views; column 4: GT part points after multi-view voting (gray: unassigned). Colors denote part names. Table: leg (red), table top (green). Chair: leg (red), cushion (green), armrest (blue). Lamp: base (red), shade (green), post (blue), rack (purple).}
\label{fig:grounding}
\end{figure}
\subsection{Grounding Part Names in Target Shapes}
\label{sec:grounding}

The part-semantic reward requires GT points for each part name, but GT meshes carry no part annotations. We locate parts with SAM~3~\cite{carion2025sam3}, prompt-tuned on data generated by MeshCoder itself, and lift its multi-view masks to 3D by voting.

\paragraph{Prompt Tuning with MeshCoder Outputs.}
We run MeshCoder on its synthetic shapes and execute the resulting programs block by block. Because every block produces its own object, rendering gives pixel-accurate part instance maps without manual labeling. For each shape, we render RGB images and instance maps from six views, three in each hemisphere, with a uniform gray material. A view is a positive sample for part name $n$ if the part is visible, and a negative sample otherwise. With all SAM~3 parameters frozen, we learn for each (category, part name) an offset $\phi_n$ added to the text token embeddings of the name, $\tilde e_n=e(n)+\phi_n$, using SAM~3's original detection and segmentation losses; negative views supervise only the absence of the part. At inference, queries whose score, multiplied by the image-level presence score, exceeds a threshold are kept, and their masks are merged into the part mask $\mathcal K_{v,n}$; if none is kept, the part is predicted absent from view $v$.

\paragraph{Rendering and Segmenting GT Meshes.}
For each training shape, we render the GT mesh from a set of views $\mathcal V$ with the same renderer and view selection, and segment every view with the tuned prompts $\tilde e_n$ for all part names in the category's vocabulary, obtaining masks $\{\mathcal K_{v,n}\}$.

\paragraph{Multi-View Voting.}
We sample points $\mathcal Q$ uniformly on the GT mesh surface, each with a face normal $\mathbf n_q$. A view $v$ validly observes a point $q$ if $q$ projects inside the image, is not occluded along the camera ray (within tolerance $\delta\cdot\operatorname{diag}(\mathcal X)$), and is not viewed at a grazing angle, $|\cos\angle(\mathbf n_q,\mathbf r_v(q))|\ge c_{\min}$. Let $o_v(q)=\mathbf{1}[v\text{ validly observes }q]$, and let $m_{v,n}(q)=\mathbf{1}[\Pi_v(q)\in\mathcal K_{v,n}]$ indicate that $q$ falls inside the mask of part $n$, where $\Pi_v$ is the projection into view $v$. For each part name $n$, we count the valid observations and the positive votes,
\begin{equation}
\begin{aligned}
\text{obs}_n(q)&=\textstyle\sum_{v\in\mathcal V}o_v(q),\\
\text{pos}_n(q)&=\textstyle\sum_{v\in\mathcal V}o_v(q)\,m_{v,n}(q).
\end{aligned}
\label{eq:votes}
\end{equation}
We define the GT part points as
\begin{equation}
\mathcal X^*_n=\big\{q\in\mathcal Q:\ \text{pos}_n(q)\ge\eta_n(q)\big\},
\label{eq:gt_parts}
\end{equation}
where $\eta_n(q)=\max\big(N_{\min},\,\tau\,\text{obs}_n(q)\big)$ is the vote threshold.
Requiring agreement from at least $N_{\min}$ views, and a majority of the views that actually see the point, suppresses segmentation errors in individual views. Fig.~\ref{fig:grounding} shows examples of the resulting GT part points.

%% ============================================================
%% Section 4: Experiments
%% Requires in the preamble:
%%   \usepackage{graphicx}
%%   \usepackage{booktabs}
%%   \usepackage{multirow}
%%   \usepackage[export]{adjustbox}   % valign=c for figure grids
%% Qualitative renders are expected under figures/qualitative/<Dataset>/<Method>/
%% (same layout as qualitative_samples.zip).
%% ============================================================

%% ---------- helpers for the qualitative figure ----------
\newcommand{\qlabel}[1]{\rotatebox[origin=c]{90}{\scriptsize #1}}
\newlength{\qwA}\setlength{\qwA}{0.093\textwidth}   % Fig. main: 10 columns, two-column width
\newlength{\qgap}\setlength{\qgap}{4pt}            % gap between dataset groups
\newcommand{\qimgA}[1]{\includegraphics[width=\qwA,valign=c]{#1}}
%% execution failure: empty cell with a small cross
\newcommand{\qimgAfail}{\parbox[c][\qwA][c]{\qwA}{\centering\scriptsize$\times$}}
%% method not evaluated on this dataset: empty cell with a dash
\newcommand{\qimgAna}{\parbox[c][\qwA][c]{\qwA}{\centering\scriptsize--}}

\section{Experiments}
\label{sec:experiments}

\subsection{Experimental Setup}
\label{sec:setup}

We build on MeshCoder~\cite{dai2025meshcoder}, a 1B model trained on procedurally generated data to generate object programs from point clouds. Other existing models that take point clouds as input perform poorly on this task without fine-tuning, so all RL methods below are applied to MeshCoder.

\paragraph{Datasets.}
We experiment on three datasets.
\textit{ShapeNetCore}~\cite{chang2015shapenet}: we use 13 categories and randomly select 3,249 shapes for RL training and 2,862 shapes for testing.
\textit{ABO}~\cite{collins2022abo}: we randomly select 721 shapes for training and 650 for testing from five categories (chair, lamp, shelf, sofa, table).
\textit{HSSD}~\cite{khanna2024hssd}: we randomly select 878 shapes from the same 13 categories as ShapeNetCore and use them only for testing models trained on ShapeNetCore, to evaluate cross-dataset generalization.

\paragraph{Methods in comparison.}
Besides MeshCoder itself, we compare with three RL baselines built on MeshCoder and with a primitive-based method.
\textit{Shape-level reward} fine-tunes MeshCoder with GRPO using only $R_{\text{shape}}$, assigned to all tokens of a rollout; this is the common practice in RL for code generation, \eg, for CAD code~\cite{kolodiazhnyi2025cadrille}.
\textit{RFT} (rejection-sampling fine-tuning) samples programs from MeshCoder and fine-tunes it on those with the highest shape-level reward.
\textit{SPO}~\cite{guo2025spo} is a segment-level RL method for language-model reasoning. We transfer it to MeshCoder by cutting segments at part-block boundaries and estimating segment advantages with Monte Carlo rollouts scored by the shape-level reward.
\textit{PrimitiveAnything}~\cite{ye2025primitiveanything} generates primitive assemblies from shapes with an autoregressive transformer trained on human-crafted abstractions; it predicts primitives directly rather than executable programs.

\paragraph{Training details.}
All training runs on a single NVIDIA RTX 6000 Ada Generation GPU. Each training step samples 4 point clouds and 8 rollouts per point cloud, \ie, 32 rollouts per step. We train for 1,000 steps on ShapeNetCore and 500 steps on ABO, with a learning rate of $1\times10^{-5}$. In all experiments, we set $\lambda_1=\lambda_2=0.35$, $\lambda_c=0.3$, $\rho=0.9$, $c_{\min}=0.2$ and $N_{\min}=2$. We also validate part grounding (Appendix~\ref{supp:grounding}). On held-out MeshCoder outputs, the tuned prompts raise the part segmentation mIoU of SAM~3 from 33.5\% to 78.7\%. On 667 ShapeNetCore training shapes annotated in PartNet~\cite{mo2019partnet}, the one-sided squared Chamfer distance from the GT part points to the annotated parts is $0.351\times10^{-2}$, 53\% lower than with direct per-view back-projection of the masks ($0.750\times10^{-2}$).

\paragraph{Metrics.}
We evaluate reconstruction accuracy with the L2 Chamfer distance (CD) and the absolute normal consistency (ANC) between points sampled from the generated and GT meshes, and execution reliability with the success rate, the fraction of test inputs whose generated program executes and yields a valid mesh. CD and ANC are averaged over successful outputs.

%% ---------- Table: ShapeNetCore + ABO ----------
\begin{table*}[t]
\centering
\caption{Reconstruction results on ShapeNetCore and ABO. CD is scaled by $10^{2}$ and averaged, like ANC, over successfully executed shapes; success rate is in \%. \textit{All}: average over all test shapes. Best results are in bold; PrimitiveAnything always yields an output and is excluded when marking the best success rate.}
\label{tab:main}
\small
\setlength{\tabcolsep}{4.5pt}
\begin{tabular}{l cccc cccc cccc}
\toprule
\multirow{2}{*}{Method} & \multicolumn{4}{c}{CD $\downarrow$} & \multicolumn{4}{c}{Absolute Normal Consistency $\uparrow$} & \multicolumn{4}{c}{Success Rate $\uparrow$} \\
\cmidrule(lr){2-5}\cmidrule(lr){6-9}\cmidrule(lr){10-13}
 & Table & Chair & Lamp & All & Table & Chair & Lamp & All & Table & Chair & Lamp & All \\
\midrule
\multicolumn{13}{c}{\textit{ShapeNetCore}} \\
\midrule
PrimitiveAnything~\cite{ye2025primitiveanything} & 20.32 & 7.70 & 13.21 & 10.46 & 0.745 & 0.732 & 0.642 & 0.743 & 100.0 & 100.0 & 100.0 & 100.0 \\
MeshCoder~\cite{dai2025meshcoder} & 2.40 & 2.25 & 2.63 & 2.64 & 0.867 & 0.774 & 0.791 & 0.815 & 98.9 & 87.0 & 93.5 & 89.2 \\
\quad + RFT & 2.06 & 2.13 & 3.23 & 2.54 & 0.869 & 0.775 & 0.791 & 0.817 & 98.9 & 87.9 & 94.4 & 90.6 \\
\quad + SPO & 2.20 & 2.30 & 2.51 & 2.05 & 0.884 & 0.775 & \textbf{0.795} & 0.820 & 99.7 & 92.5 & 99.1 & 96.5 \\
\quad + shape-level reward & 1.94 & 2.29 & 2.27 & 2.27 & 0.884 & 0.784 & 0.788 & 0.826 & 99.2 & 92.2 & 97.2 & 96.5 \\
\quad + \textbf{PartPO} & \textbf{0.85} & \textbf{1.57} & \textbf{1.92} & \textbf{1.24} & \textbf{0.890} & \textbf{0.790} & 0.788 & \textbf{0.833} & \textbf{99.9} & \textbf{93.1} & \textbf{100.0} & \textbf{97.8} \\
\midrule
\multicolumn{13}{c}{\textit{ABO}} \\
\midrule
PrimitiveAnything~\cite{ye2025primitiveanything} & 3.81 & 1.91 & 5.79 & 2.76 & 0.802 & 0.796 & 0.720 & 0.780 & 100.0 & 100.0 & 100.0 & 100.0 \\
MeshCoder~\cite{dai2025meshcoder} & 3.54 & 1.22 & 3.91 & 2.37 & 0.832 & 0.810 & 0.778 & 0.797 & \textbf{100.0} & 79.3 & 96.8 & 58.6 \\
\quad + RFT & 3.68 & 1.26 & 3.88 & 2.36 & 0.835 & 0.806 & 0.781 & 0.797 & \textbf{100.0} & 78.7 & 97.6 & 58.5 \\
\quad + SPO & 3.05 & 1.65 & 2.02 & 1.30 & 0.844 & 0.798 & 0.841 & 0.814 & \textbf{100.0} & 91.3 & 99.2 & 97.5 \\
\quad + shape-level reward & 3.59 & 1.34 & 2.53 & 1.38 & 0.828 & 0.804 & 0.821 & 0.816 & \textbf{100.0} & \textbf{92.7} & \textbf{100.0} & 90.0 \\
\quad + \textbf{PartPO} & \textbf{2.47} & \textbf{1.07} & \textbf{1.49} & \textbf{0.95} & \textbf{0.848} & \textbf{0.821} & \textbf{0.853} & \textbf{0.831} & \textbf{100.0} & 92.0 & \textbf{100.0} & \textbf{97.8} \\
\bottomrule
\end{tabular}
\end{table*}

%% ---------- Figure: ABO + ShapeNetCore + HSSD ----------
%% Columns -- ABO: armchair, lamp, square table, round table;
%%            ShapeNetCore: chair, floor lamp, sofa, bookshelf;
%%            HSSD (zero-shot): round table, console table.
\begin{figure*}[t]
\centering
\setlength{\tabcolsep}{0pt}
\renewcommand{\arraystretch}{0}
\begin{tabular}{c cccc@{\hspace{\qgap}} cccc@{\hspace{\qgap}} cc}
 & \multicolumn{4}{c}{\small ABO} & \multicolumn{4}{c}{\small ShapeNetCore} & \multicolumn{2}{c}{\small HSSD (zero-shot)} \\[3pt]
\qlabel{GT} &
  \qimgA{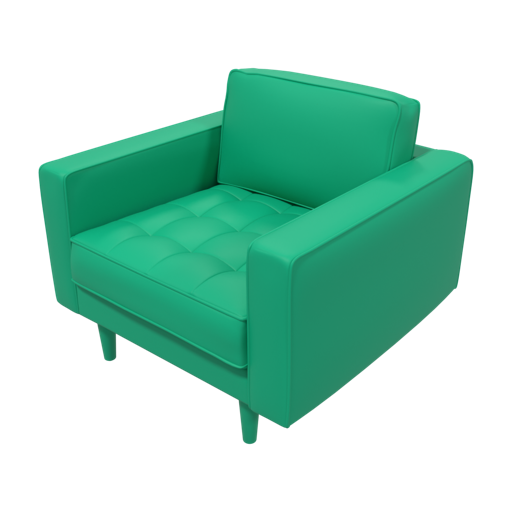} &
  \qimgA{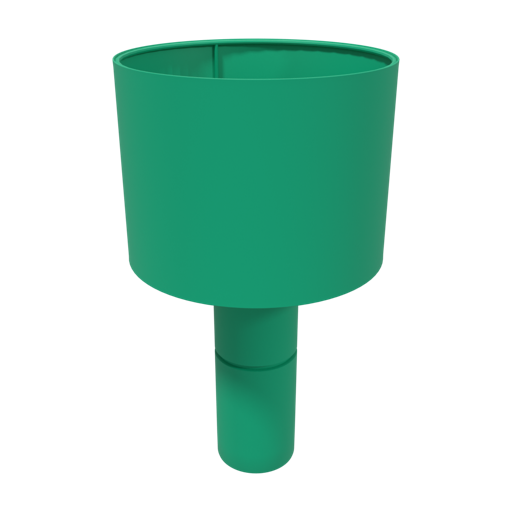} &
  \qimgA{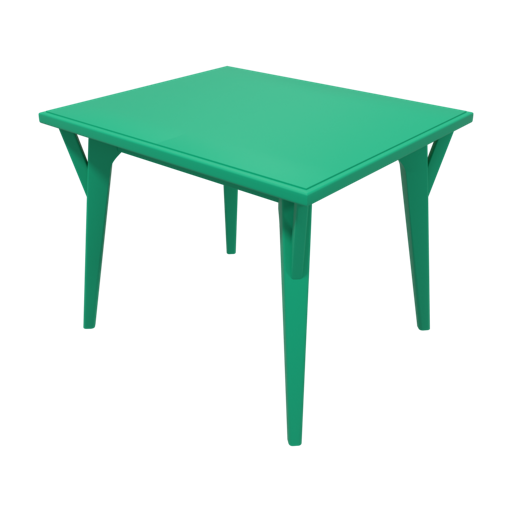} &
  \qimgA{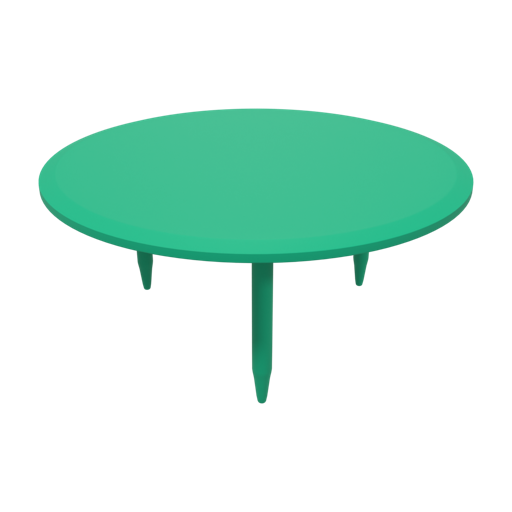} &
  \qimgA{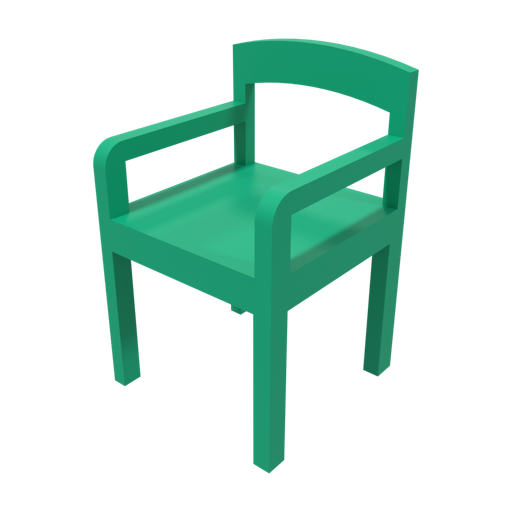} &
  \qimgA{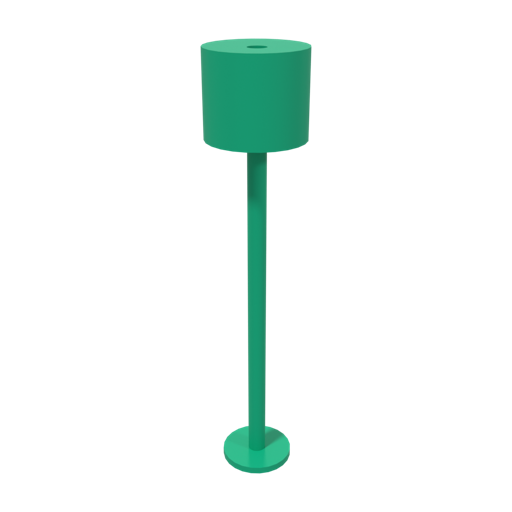} &
  \qimgA{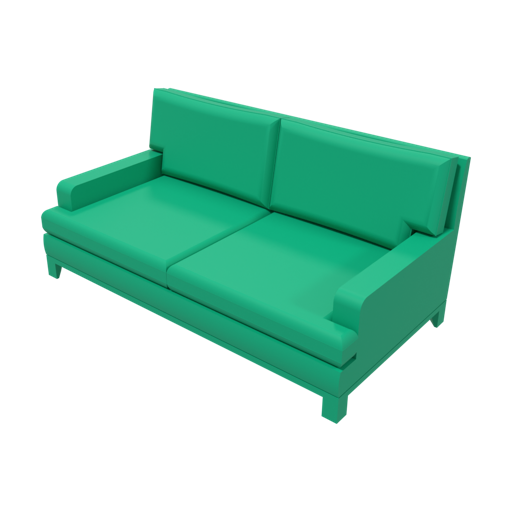} &
  \qimgA{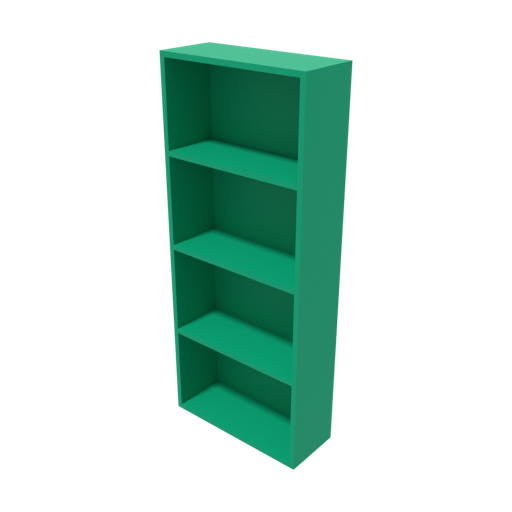} &
  \qimgA{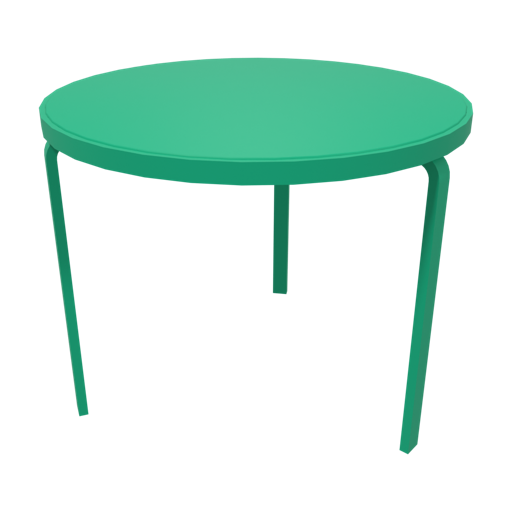} &
  \qimgA{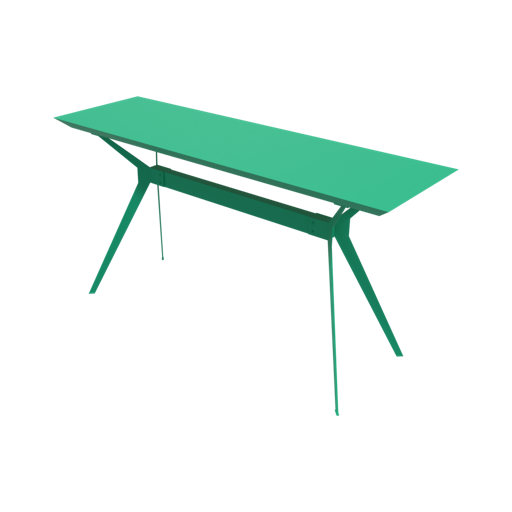} \\
\qlabel{MeshCoder} &
  \qimgAfail &
  \qimgA{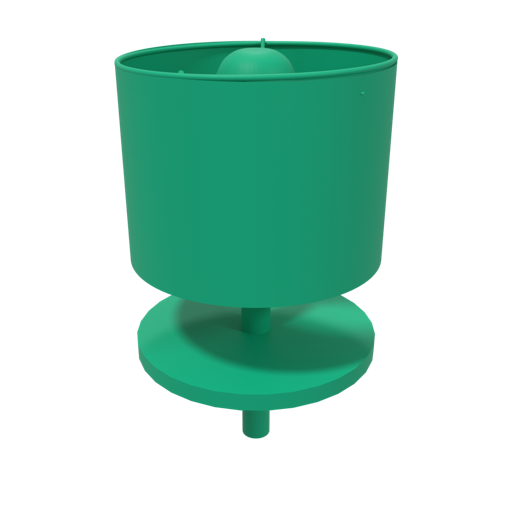} &
  \qimgA{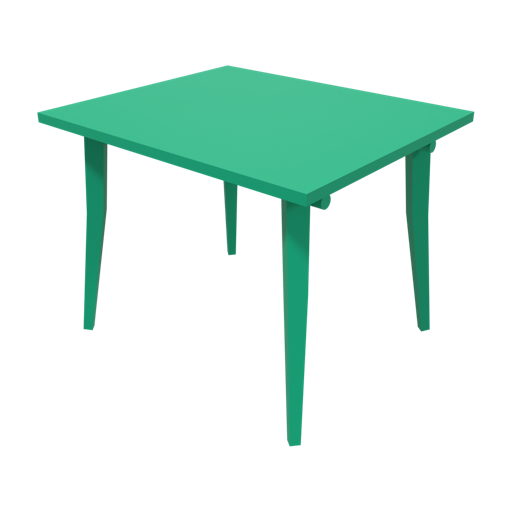} &
  \qimgA{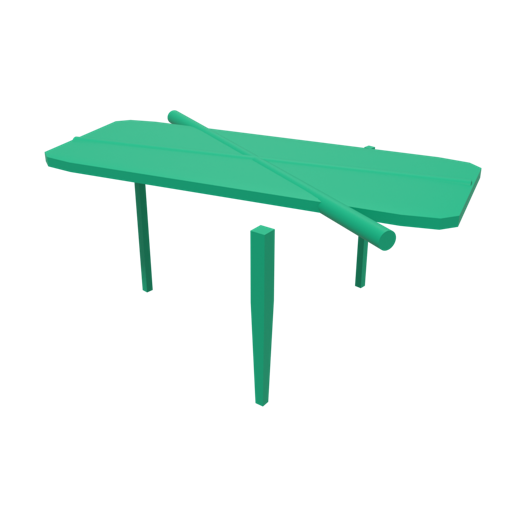} &
  \qimgA{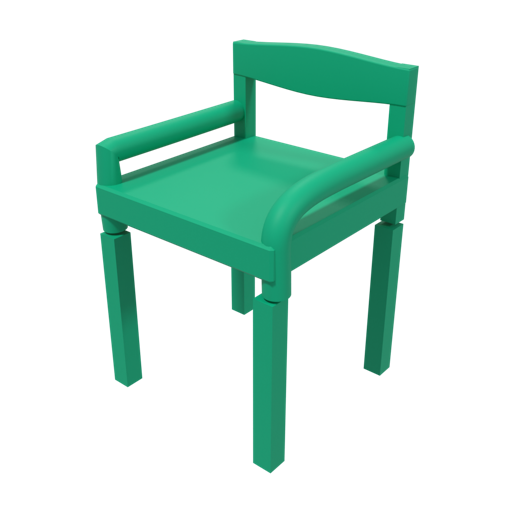} &
  \qimgAfail &
  \qimgA{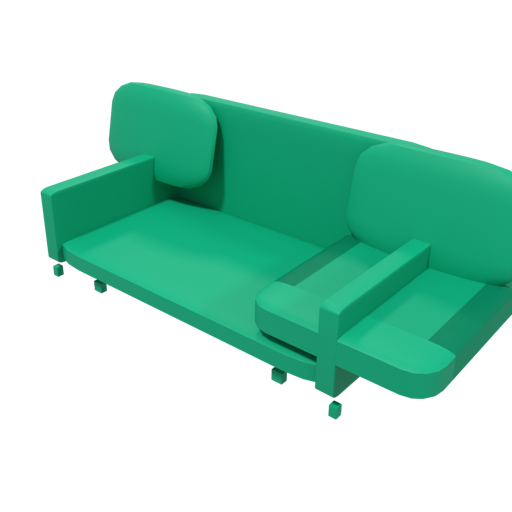} &
  \qimgA{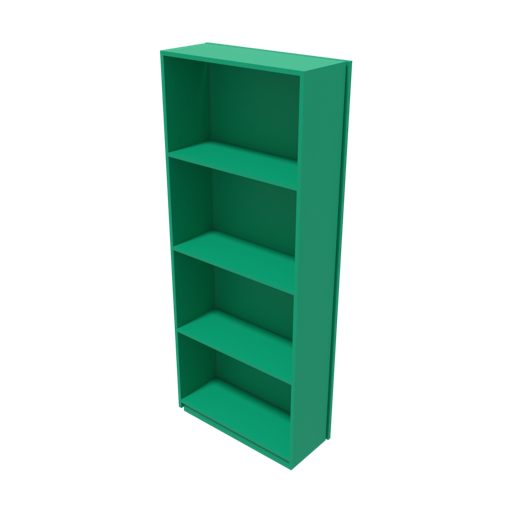} &
  \qimgA{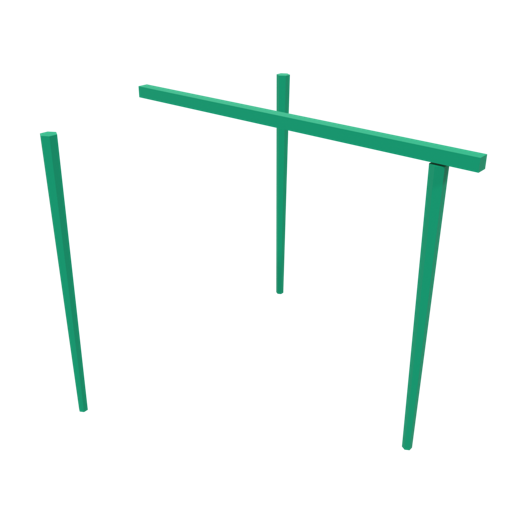} &
  \qimgA{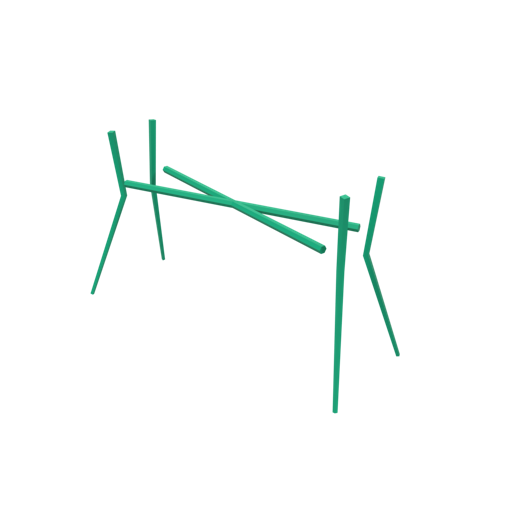} \\
\qlabel{+ SPO} &
  \qimgA{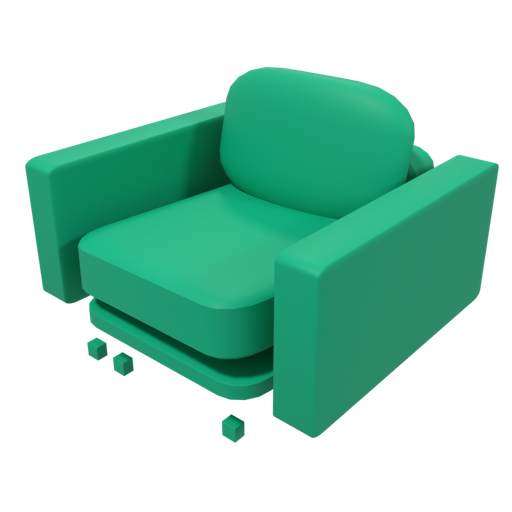} &
  \qimgA{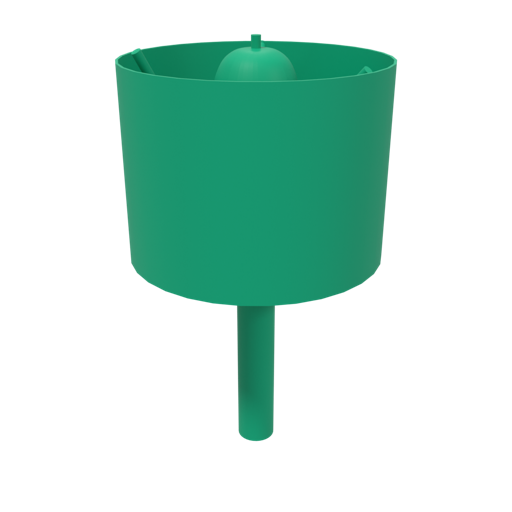} &
  \qimgA{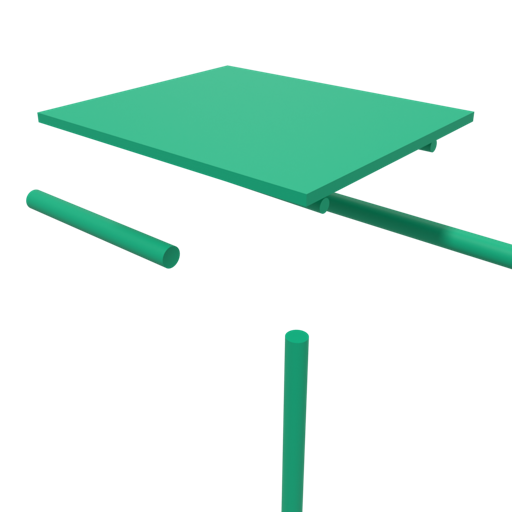} &
  \qimgA{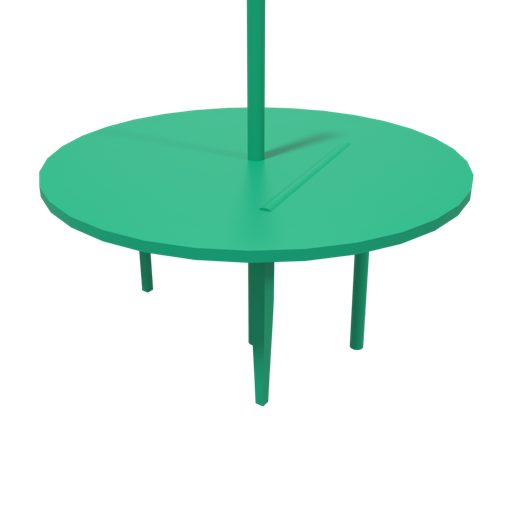} &
  \qimgA{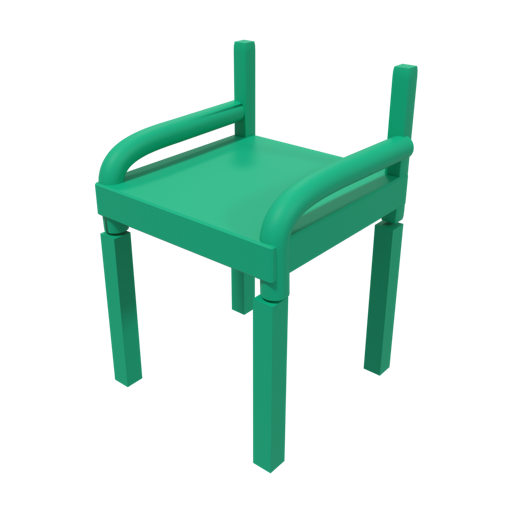} &
  \qimgA{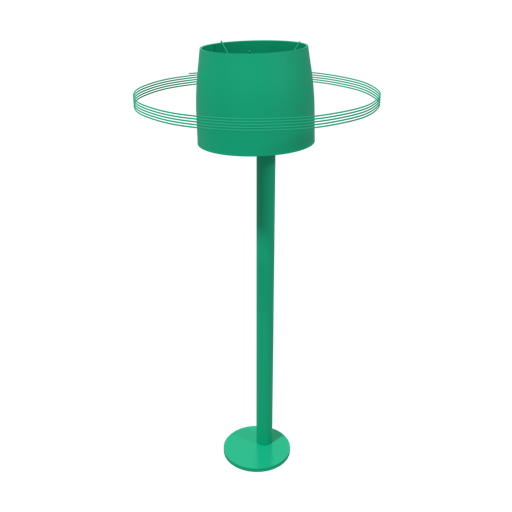} &
  \qimgA{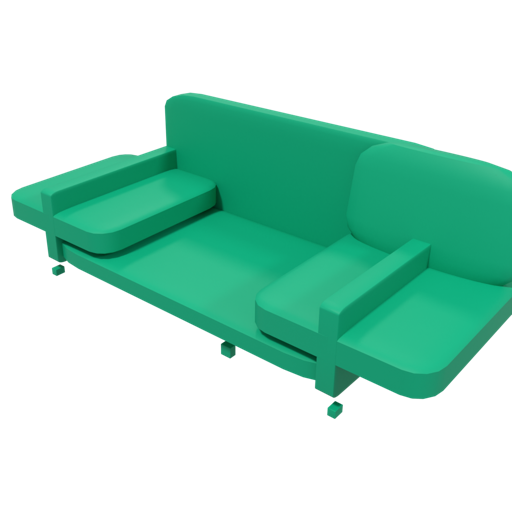} &
  \qimgA{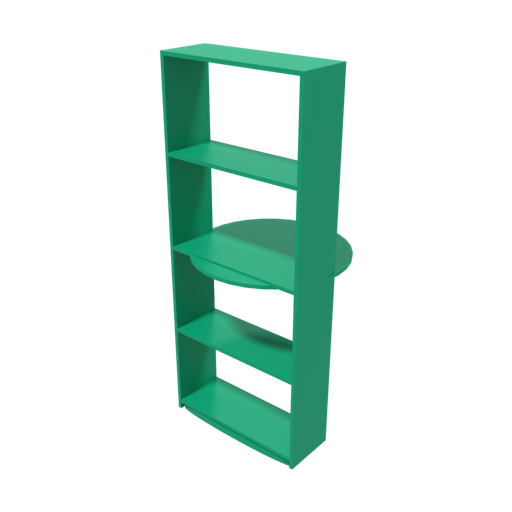} &
  \qimgAna &
  \qimgAna \\
\qlabel{+ Shape-level} &
  \qimgA{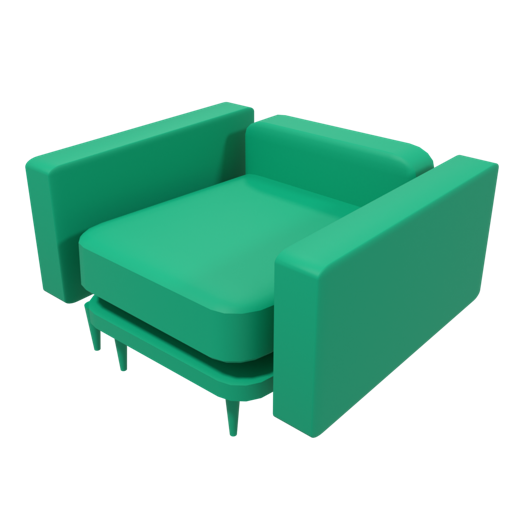} &
  \qimgA{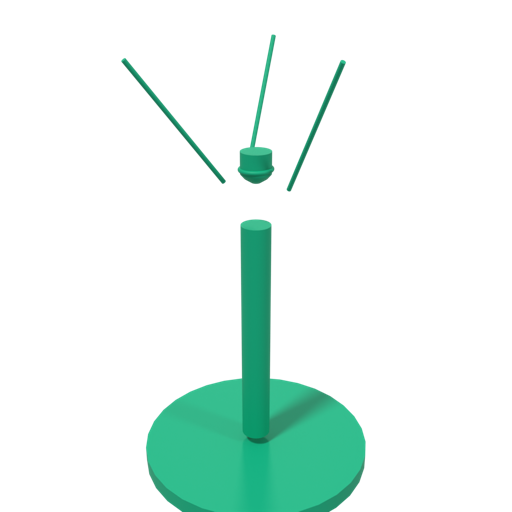} &
  \qimgA{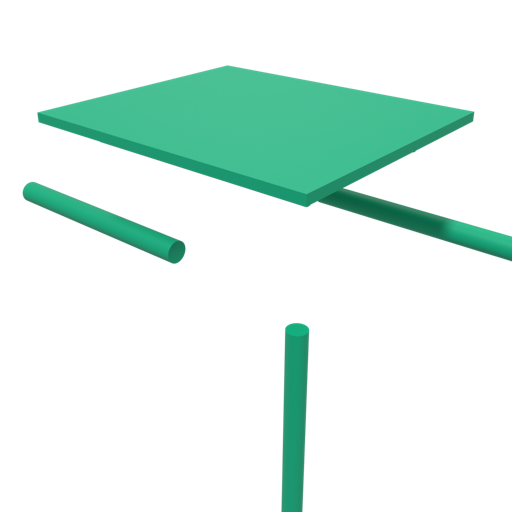} &
  \qimgA{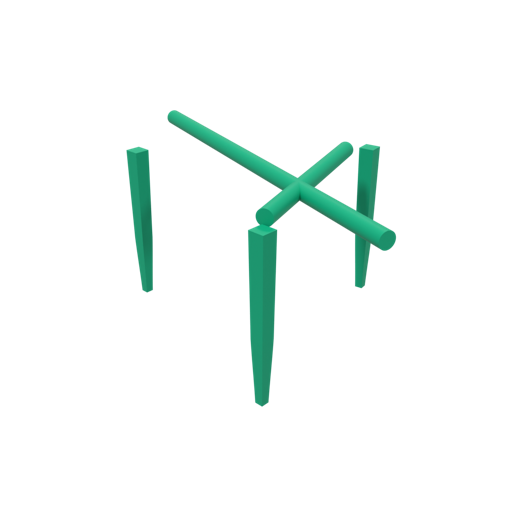} &
  \qimgA{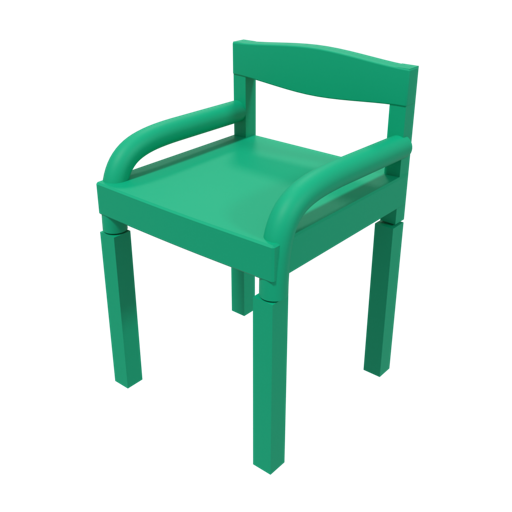} &
  \qimgA{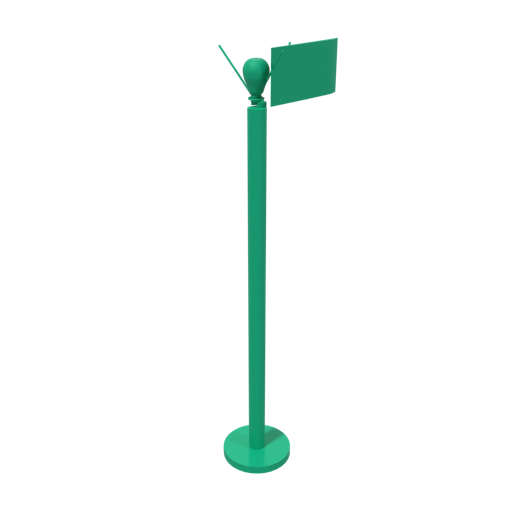} &
  \qimgA{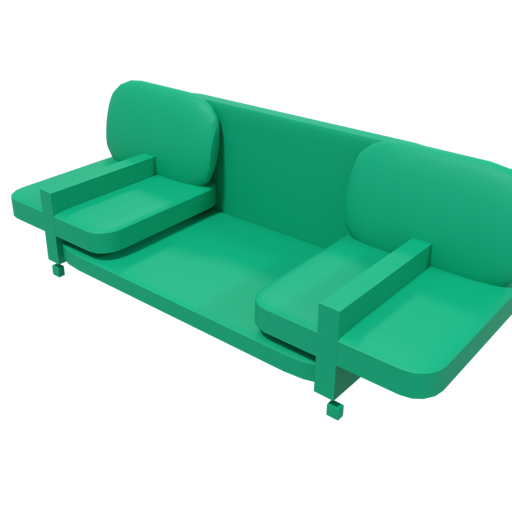} &
  \qimgA{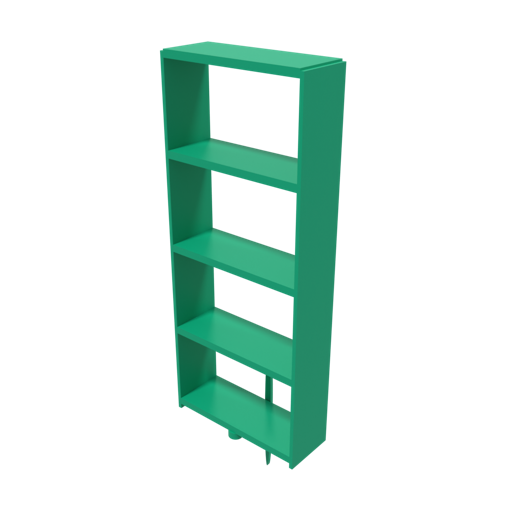} &
  \qimgA{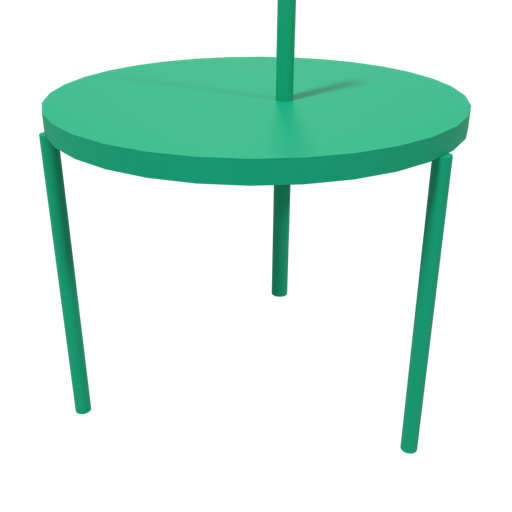} &
  \qimgA{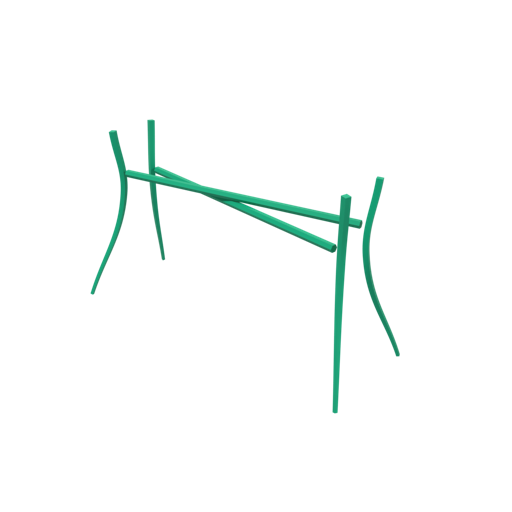} \\
\qlabel{+ \textbf{PartPO}} &
  \qimgA{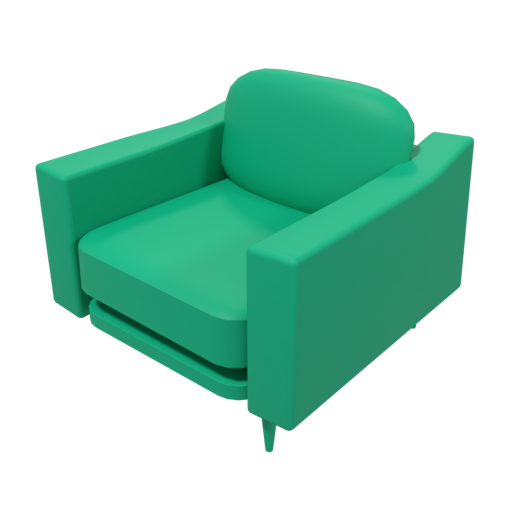} &
  \qimgA{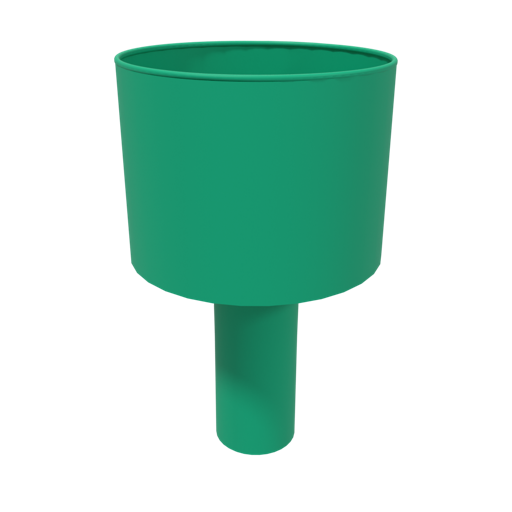} &
  \qimgA{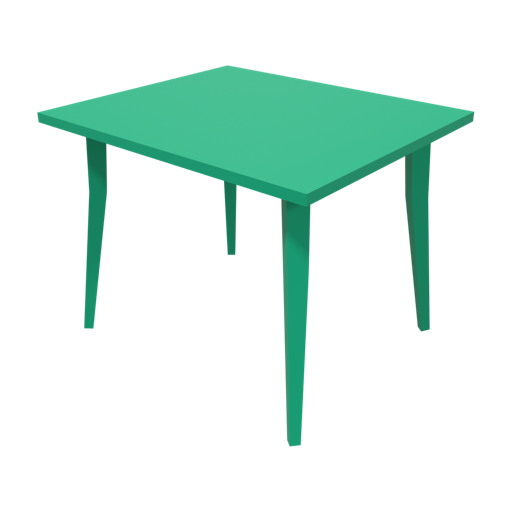} &
  \qimgA{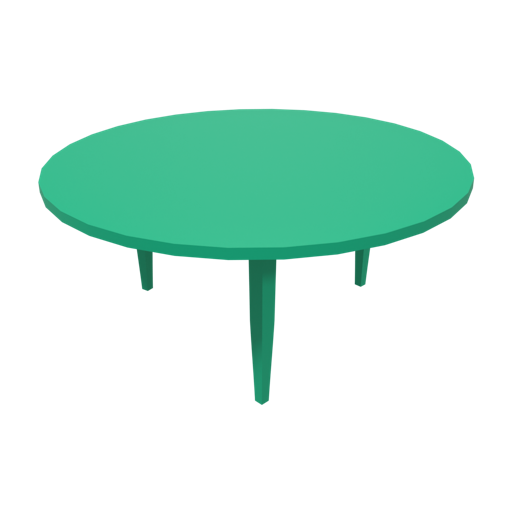} &
  \qimgA{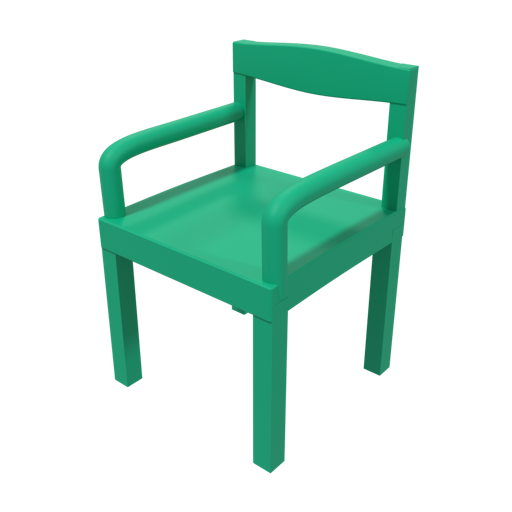} &
  \qimgA{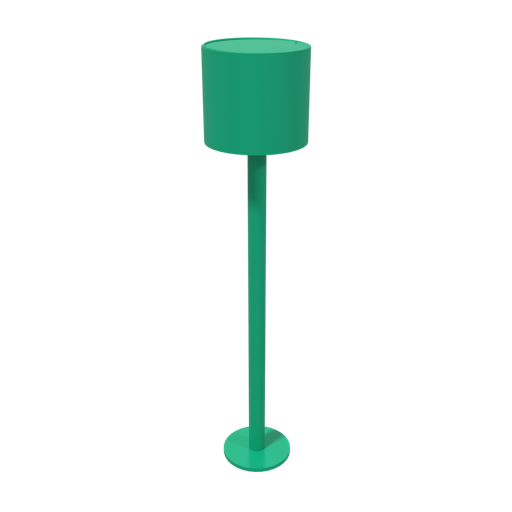} &
  \qimgA{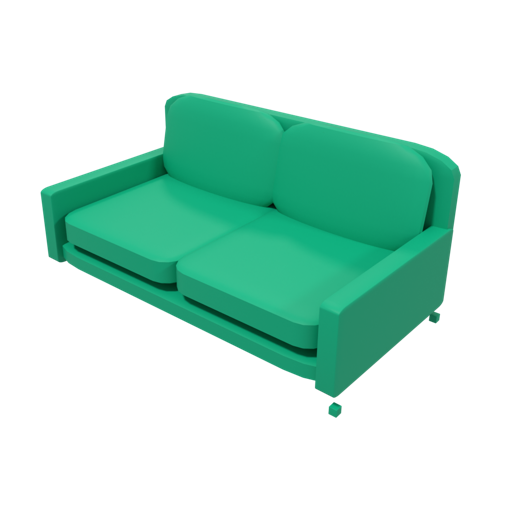} &
  \qimgA{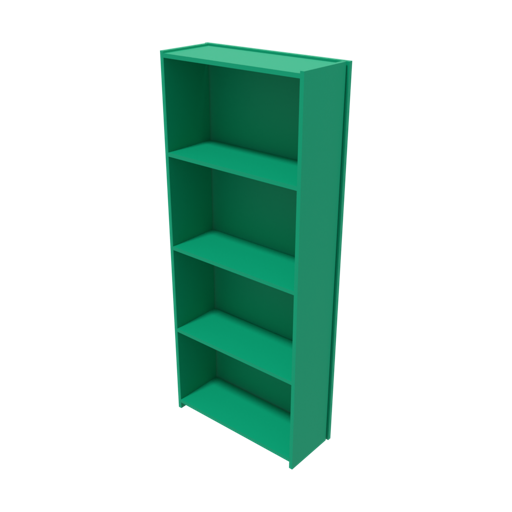} &
  \qimgA{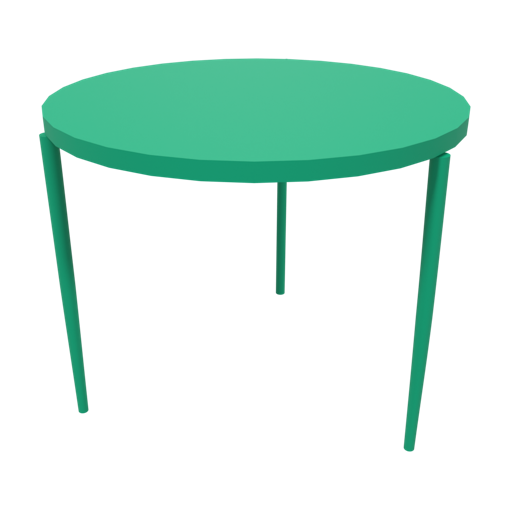} &
  \qimgA{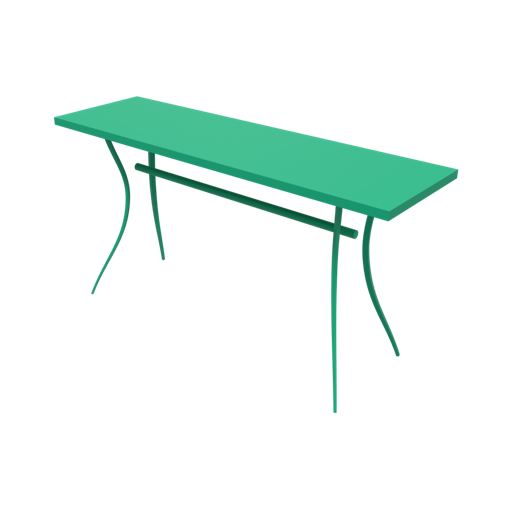} \\
\end{tabular}
\caption{Qualitative comparison on ABO (left), ShapeNetCore (middle) and HSSD (right; models trained on ShapeNetCore, tested without further training). $\times$ marks programs that fail to execute; SPO is not evaluated on HSSD (--). PartPO reconstructs each part in place, while the baselines produce missing, floating or misplaced parts.}
\label{fig:qual_main}
\end{figure*}

\subsection{Main Results}
\label{sec:main_results}

\paragraph{Results on ShapeNetCore.}
We report scores in Tab.~\ref{tab:main} (top) and qualitative results in Fig.~\ref{fig:qual_main} (middle). PartPO achieves the best overall CD, ANC and success rate, more than halving the CD of MeshCoder (2.64 to 1.24). SPO and the shape-level reward improve over MeshCoder by smaller margins, while RFT brings only marginal gains. The qualitative results show where the difference comes from: the baselines usually recover the overall shape but get individual parts wrong, such as a missing backrest rail, misplaced sofa armrests, a malformed lampshade, or extra geometry on the lamp and the bookshelf, whereas PartPO reconstructs each part in place. PrimitiveAnything always yields an output, but its primitive assemblies are far less accurate.

\paragraph{Results on ABO.}
Scores are in Tab.~\ref{tab:main} (bottom) and qualitative results in Fig.~\ref{fig:qual_main} (left). The domain gap is larger on ABO: MeshCoder executes successfully on only 58.6\% of the test shapes, and RFT leaves this nearly unchanged. PartPO raises the success rate to 97.8\% and reduces the CD by 60\%, with the best CD and ANC on every listed category. SPO nearly matches its success rate but not its accuracy, and the shape-level reward falls behind on both. In Fig.~\ref{fig:qual_main}, MeshCoder fails to execute on the armchair; SPO and the shape-level baseline produce valid programs, but with detached legs and stretchers, a missing tabletop or lampshade, or stray fragments, whereas PartPO yields complete and well-assembled shapes.

\paragraph{Accuracy on commonly executed shapes.}
Because CD and ANC are averaged over the successful outputs of each method, a method could appear more accurate simply by failing on harder shapes. We therefore also evaluate on the shapes that all program-generating methods execute successfully (2,505 on ShapeNetCore and 362 on ABO). PartPO still achieves the lowest CD, 1.03 and 1.29, compared with at best 1.84 and 1.73 for the baselines, as well as the highest ANC (Appendix~\ref{supp:intersection}).

\paragraph{Results on HSSD.}
To test cross-dataset generalization, we evaluate the models trained on ShapeNetCore on HSSD without further training (Tab.~\ref{tab:hssd}, Fig.~\ref{fig:qual_main} right). PartPO improves both CD and success rate over MeshCoder and the shape-level reward, with comparable ANC. In Fig.~\ref{fig:qual_main}, MeshCoder omits both tabletops, and the shape-level baseline lets a leg pierce the round tabletop and still omits the console tabletop; PartPO recovers both.

%% ---------- Table: HSSD ----------
\begin{table}[t]
\centering
\caption{Zero-shot results on HSSD of models trained on ShapeNetCore. CD is scaled by $10^{2}$ and success rate is in \%.}
\label{tab:hssd}
\small
\begin{tabular}{l ccc}
\toprule
Method & CD $\downarrow$ & ANC $\uparrow$ & Success $\uparrow$ \\
\midrule
MeshCoder~\cite{dai2025meshcoder} & 3.12 & 0.809 & 81.3 \\
\quad + shape-level reward & 2.92 & \textbf{0.817} & 88.3 \\
\quad + \textbf{PartPO} & \textbf{2.69} & 0.816 & \textbf{89.2} \\
\bottomrule
\end{tabular}
\end{table}

\subsection{Part-Level Evaluation}
\label{sec:part_eval}

%% ---------- Table: part-level evaluation ----------
\begin{table}[t]
\centering
\caption{Part-level comparison between PartPO and baselines on the 504 ShapeNetCore test shapes (table, chair and lamp) annotated in PartNet. Part CD is scaled by $10^{3}$; success rate, recall and precision are in \%. Best results are in bold.}
\label{tab:part}
\resizebox{\columnwidth}{!}{%
\setlength{\tabcolsep}{3pt}
\begin{tabular}{l ccccc}
\toprule
Method & Success $\uparrow$ & Part CD $\downarrow$ & Part NC $\uparrow$ & Recall $\uparrow$ & Precision $\uparrow$ \\
\midrule
MeshCoder~\cite{dai2025meshcoder} & 94.4 & 6.72 & 0.626 & 72.7 & 87.4 \\
\quad + shape-level reward & 94.4 & 6.10 & 0.627 & 71.8 & 88.4 \\
\quad + \textbf{PartPO} & \textbf{96.4} & \textbf{4.74} & \textbf{0.640} & \textbf{74.2} & \textbf{89.2} \\
\bottomrule
\end{tabular}}
\end{table}

\paragraph{Setup.}
Shape-level metrics do not show whether a program decomposes the object into the right parts. We therefore evaluate parts on the 504 ShapeNetCore test shapes of tables, chairs and lamps that are annotated in PartNet~\cite{mo2019partnet}; these annotations are not used in training. In each program, code blocks with the same part name are merged into one predicted part. PartNet organizes the parts of each shape in a tree, where a part such as a chair back may be split into finer sub-parts. We use every node of the tree except the root as a GT part, together with the atomic meshes, the smallest annotated pieces, so that a predicted part can match either a whole back or one of its sub-parts. We then match predicted and GT parts greedily in order of increasing Chamfer distance. A pair is accepted only if its distance is below a threshold, the predicted part is not yet matched, and the GT part does not overlap an already matched one, such as a sub-part of a matched back. Thus each predicted part matches at most one GT part, and no surface region is counted twice. Part CD and Part NC are averaged over the matched pairs. Recall is the fraction of the object's surface area covered by matched GT parts, so it drops when parts are missing. Precision is the fraction of predicted parts that are matched, so it drops when parts are extra or misplaced. Details are in Appendix~\ref{supp:part_eval}.

\paragraph{Results.}
Tab.~\ref{tab:part} shows that PartPO improves every part-level metric. It reduces the Part CD of MeshCoder from 6.72 to 4.74 and raises Part NC from 0.626 to 0.640, so the parts it builds are geometrically more accurate. It also raises recall from 72.7\% to 74.2\% and precision from 87.4\% to 89.2\%, so its programs miss fewer parts and produce fewer parts that match nothing, while executing successfully more often (96.4\% \vs 94.4\%). The shape-level reward, in contrast, improves Part CD and precision only modestly (6.10 and 88.4\%) and lowers recall to 71.8\%: optimizing whole-shape similarity alone does not make the program build the right parts. The improvements of PartPO in Sec.~\ref{sec:main_results} therefore come from better individual parts, which is what its part-level credit assignment targets.

\subsection{Ablation Studies}
\label{sec:ablation}

%% ---------- Table: ablation on ABO (lamp, table) ----------
\begin{table}[t]
\centering
\caption{Ablation on ABO (lamp and table). Scalar sum adds the three rewards into one scalar assigned to all tokens; PartPO applies each reward only to the tokens it evaluates. CD is scaled by 10² and success rate is in \%.}
\label{tab:ablation}
\resizebox{\columnwidth}{!}{%
\setlength{\tabcolsep}{3pt}
\begin{tabular}{l ccc ccc ccc}
\toprule
\multirow{2}{*}{Method} & \multicolumn{3}{c}{CD $\downarrow$} & \multicolumn{3}{c}{Abs. Normal Consistency $\uparrow$} & \multicolumn{3}{c}{Success Rate $\uparrow$} \\
\cmidrule(lr){2-4}\cmidrule(lr){5-7}\cmidrule(lr){8-10}
 & Lamp & Table & All & Lamp & Table & All & Lamp & Table & All \\
\midrule
MeshCoder~\cite{dai2025meshcoder} & 4.70 & 4.72 & 4.71 & 0.759 & 0.709 & 0.728 & 93.4 & 95.3 & 94.5 \\
\quad + shape-level & 3.07 & 4.93 & 4.19 & 0.787 & 0.728 & 0.751 & 90.7 & 88.9 & 89.6 \\
\quad + shape-level + part-level & 2.70 & 3.39 & 3.13 & 0.812 & 0.729 & 0.760 & \textbf{99.3} & 98.3 & 98.7 \\
\quad + scalar sum of all rewards & 2.37 & 4.45 & 3.63 & 0.803 & 0.717 & 0.751 & 94.0 & 93.2 & 93.5 \\
\quad + \textbf{PartPO} & \textbf{1.86} & \textbf{3.28} & \textbf{2.72} & \textbf{0.819} & \textbf{0.739} & \textbf{0.770} & \textbf{99.3} & \textbf{99.6} & \textbf{99.5} \\
\bottomrule
\end{tabular}}
\end{table}

We ablate PartPO on two ABO categories, lamp and table, using 467 training and 385 test shapes disjoint from the main experiments, and train each RL variant for 300 steps. Starting from MeshCoder, we add the shape-level reward, then the part-level reward, and finally the category reward, either summed with the others into one scalar for all tokens or applied only to the tokens it evaluates, as in PartPO (Tab.~\ref{tab:ablation}).

The shape-level reward alone lowers the CD of MeshCoder but also its success rate, and even increases the CD on tables. Adding the part-level reward improves all summary metrics, and the category reward brings a further gain: it encourages the policy to name the object category correctly, which is the precondition for the part-level reward to be informative. Finally, the structure-aligned credit assignment is the key. Summing the same three rewards into a scalar provides the same reward information as PartPO, yet raises the CD from 2.72 to 3.63, worse than the localized variant even without the category reward, and drops the success rate below that of MeshCoder. The rewards therefore help only when each of them reaches the tokens that produced what it evaluates.

%% ============================================================
%% Section 5: Conclusion
%% ============================================================

\section{Conclusion}
\label{sec:conclusion}

This paper presents PartPO, an RL method for shape program generation that builds on the self-describing part structure of shape programs. Each reward reaches only the tokens that produce what it evaluates: part-level rewards act on part blocks, the category reward on the object name, and the shape-level reward on the whole program. On ShapeNetCore and ABO, PartPO improves the reconstruction accuracy and execution success of MeshCoder, outperforms rejection-sampling fine-tuning and trajectory- and segment-level RL, and generalizes to the unseen HSSD dataset. More broadly, our results suggest that when generated code exposes its own structure, verifying each component against the target provides dense credit without a learned reward model or extra rollouts.

\paragraph{Limitations.}
We evaluate a single base model and DSL. Few models generate part-structured programs from point clouds, and MeshCoder is the most capable openly available one. PartPO only requires executable programs with named part blocks. Verifying its transfer to other shape program generators is future work.
\section*{LLM Usage Statement}
\label{sec:llm}
We used Claude Sonnet 5.5 (Anthropic) in the preparation of this paper in
the following limited ways. (1) Writing: the model was used to polish the
wording and rhetoric of author-written text. All suggestions were reviewed,
edited, and approved by the authors. (2) Figures: the model provided advice
on the layout and visual design of the pipeline diagram; the figure itself
was created by the authors. (3) Code: the model was used to assist in debugging our implementation of
the 2D-to-3D lifting module, e.g., diagnosing error messages and suggesting
fixes for specific implementation bugs. The lifting method itself was
designed by the authors, the code was written by the authors, and every
suggested change was reviewed, tested, and verified by the authors before
being adopted. The authors take full responsibility for all content of this paper.
{
    \small
    \bibliographystyle{ieeenat_fullname}
    \bibliography{human}
}
\clearpage
\appendix
%% ============================================================
%% Supplementary Material
%% Requires in the preamble:
%%   \usepackage{amsmath,amssymb,booktabs}
%%   \usepackage{algorithm}
%%   \usepackage{algpseudocode}
%% In the CVPR template, place this after \maketitlesupplementary.
%% ============================================================
\section{Part-Level Evaluation Details}
\label{supp:part_eval}

This section details the part-level evaluation in Sec.~4.3. Shape-level metrics such as the overall Chamfer distance measure how close a reconstruction is to the target, but not whether the generated program decomposes the object into semantically meaningful parts. We therefore match the parts produced by each program with the hierarchical part annotations of PartNet~\cite{mo2019partnet} and measure the geometric accuracy, coverage and correctness of the matched parts. The evaluation uses the 504 test shapes of ShapeNetCore in the table, chair and lamp categories that are annotated in PartNet.

\subsection{Predicted Parts}
Each block of a generated program carries a part name (\eg, \texttt{leg}, \texttt{seat}) and produces a part mesh when executed. Because one semantic part is often built by several blocks, such as the four legs of a chair, we normalize the part names and merge the blocks with the same normalized name into one predicted part. The predicted parts of an object are denoted by $\mathcal P=\{P_1,\dots,P_M\}$, each a triangle mesh. A program that fails to execute or produces no part mesh counts as an execution failure.

\subsection{Ground-Truth Part Candidates}
PartNet annotates each object with a part hierarchy, in which every node is a semantic part composed of one or more atomic meshes, the smallest indivisible geometric units of PartNet. For a node $g$, we denote its set of atomic meshes by $\mathcal O(g)$; the geometry of the node is their union. All nodes except the root, which is the whole object, serve as GT part candidates, so a predicted part can align with the annotation at any granularity: with a coarse part such as a chair back or with one of its finer sub-parts. In addition, each atomic mesh of a node that contains several is added as a separate candidate, to cover decompositions finer than the annotated leaves. When several nodes have identical atomic mesh sets, as in a chain of single-child nodes that differ only in name, we keep one of them. The final candidate set is $\mathcal G=\{G_1,\dots,G_K\}$.

These candidates have a key property: the atomic mesh sets of any two candidates are either disjoint or nested, and $\mathcal O(G_j)\cap\mathcal O(G_k)\neq\emptyset$ if and only if $G_j$ and $G_k$ are the same node or one is an ancestor of the other. The matching below uses this property to enforce consistency with the hierarchy.

\paragraph{Coordinate alignment.}
PartNet meshes lie in a different coordinate frame from the normalized frame used by the models. We align each PartNet mesh to the normalized GT point cloud $\mathcal X$ of the same object. For each of the 24 axis-aligned rotations $R$, we apply a scaling and translation that match the center and longest side of the rotated mesh's bounding box to those of the bounding box of $\mathcal X$, and we select the transformation with the smallest Chamfer distance to $\mathcal X$. The same transformation is applied to all GT part candidates of the object.

\subsection{Distance}
For two meshes $A$ and $B$, we sample point sets $S_A$ and $S_B$ uniformly on their surfaces with area weighting and use the bidirectional Chamfer distance with squared Euclidean distances:
\begin{equation}
\begin{split}
\mathrm{CD}(A,B)={}&\frac1{|S_A|}\sum_{a\in S_A}\min_{b\in S_B}\|a-b\|_2^2\\
&+\frac1{|S_B|}\sum_{b\in S_B}\min_{a\in S_A}\|a-b\|_2^2.
\end{split}
\label{eq:supp_cd}
\end{equation}

\subsection{Part Matching}
We compute $\mathrm{CD}(P_i,G_j)$ for all pairs of predicted parts and GT candidates and obtain a matching $\mathcal M\subseteq\mathcal P\times\mathcal G$ greedily (Algorithm~\ref{alg:matching}). The pairs are visited in ascending order of distance, and a pair $(P_i,G_j)$ is accepted if and only if
\begin{enumerate}
    \item \emph{distance threshold}: $\mathrm{CD}(P_i,G_j)\le\tau$, \ie, the two are geometrically close;
    \item \emph{one-to-one}: $P_i$ has not been matched;
    \item \emph{hierarchical exclusivity}: $\mathcal O(G_j)$ is disjoint from the atomic meshes of all matched GT candidates.
\end{enumerate}
By the property above, condition 3 is equivalent to requiring that $G_j$ is neither matched nor an ancestor or descendant of a matched node. The resulting matching is therefore one-to-one, and the matched GT candidates do not overlap, so the same region is never counted both as a coarse part and as one of its sub-parts. Each predicted part can thus align with the annotation at its most suitable granularity, while the matched GT candidates form a subset of a non-overlapping partition of the object. We use $\tau=0.05$.

\begin{algorithm}[t]
\caption{Greedy part matching with hierarchical constraints.}
\label{alg:matching}
\begin{algorithmic}[1]
\Require predicted parts $\mathcal P$, GT candidates $\mathcal G$, threshold $\tau$
\State $\mathcal M\gets\emptyset$; $\mathcal U\gets\emptyset$ \Comment{matched predicted parts}
\State $\mathcal O_{\text{used}}\gets\emptyset$ \Comment{occupied atomic meshes}
\For{$(P_i,G_j)$ in ascending order of $\mathrm{CD}(P_i,G_j)$}
    \If{$\mathrm{CD}(P_i,G_j)>\tau$} \textbf{break} \EndIf
    \If{$P_i\notin\mathcal U$ \textbf{and} $\mathcal O(G_j)\cap\mathcal O_{\text{used}}=\emptyset$}
        \State $\mathcal M\gets\mathcal M\cup\{(P_i,G_j)\}$
        \State $\mathcal U\gets\mathcal U\cup\{P_i\}$; $\mathcal O_{\text{used}}\gets\mathcal O_{\text{used}}\cup\mathcal O(G_j)$
    \EndIf
\EndFor
\State \Return $\mathcal M$
\end{algorithmic}
\end{algorithm}

\subsection{Metrics}
For each successfully executed shape, we compute the following metrics on the matching $\mathcal M$. After matching, the geometric metrics are recomputed with denser point sampling on each part for stability.

\paragraph{Part CD ($\downarrow$).}
The Chamfer distance of Eq.~\eqref{eq:supp_cd} averaged over the matched pairs, measuring the geometric accuracy of the correctly identified parts:
\begin{equation}
\text{Part-CD}=\frac1{|\mathcal M|}\sum_{(P,G)\in\mathcal M}\mathrm{CD}(P,G).
\end{equation}

\paragraph{Part NC ($\uparrow$).}
For a point $a$ sampled on mesh $A$, let $\mathbf n_A(a)$ be the normal of its triangle and $\mathbf n_B(\pi_B(a))$ the normal of the triangle containing its nearest point on mesh $B$. The one-sided normal consistency is
\begin{equation}
\mathrm{NC}(A\to B)=\frac1{|S_A|}\sum_{a\in S_A}\big|\mathbf n_A(a)\cdot\mathbf n_B(\pi_B(a))\big|,
\end{equation}
and the bidirectional one is $\mathrm{NC}(A,B)=\frac12\big(\mathrm{NC}(A\to B)+\mathrm{NC}(B\to A)\big)$. The absolute value makes the metric independent of the orientation convention of mesh faces. Part NC averages $\mathrm{NC}(P,G)$ over the matched pairs and measures the fidelity of surface orientation and local detail.

\paragraph{Recall ($\uparrow$).}
Recall measures how much of the object's surface is covered by correctly identified parts and thus reflects missing parts:
\begin{equation}
\text{Recall}=\frac{\sum_{(P,G)\in\mathcal M}\operatorname{Area}(G)}{\operatorname{Area}(\text{object})},
\end{equation}
where $\operatorname{Area}(\cdot)$ is the surface area and the denominator is the total area of all atomic meshes of the object. Because the matched GT candidates do not overlap, no area is counted twice and the ratio never exceeds 1. Weighting by area rather than by count makes the metric independent of the granularity of the annotation: covering the same region yields the same recall whether a predicted part matches a coarse part or several of its sub-parts.

\paragraph{Precision ($\uparrow$).}
Precision measures how many predicted parts correspond to real parts and thus reflects extra or misplaced parts:
\begin{equation}
\text{Precision}=\frac{\big|\{P:(P,G)\in\mathcal M\}\big|}{|\mathcal P|}.
\end{equation}

\paragraph{F1 ($\uparrow$).}
The harmonic mean of precision and recall summarizes the completeness and correctness of the decomposition:
\begin{equation}
\text{F1}=\frac{2\cdot\text{Precision}\cdot\text{Recall}}{\text{Precision}+\text{Recall}}.
\end{equation}

\subsection{Aggregation over the Dataset}
The success rate is the fraction of all test shapes whose program executes and produces part meshes. Part CD and Part NC are first averaged over the matched pairs within each shape and then over shapes; shapes without any match are excluded from these two metrics. Recall, precision and F1 are averaged over successfully executed shapes. Since F1 is computed per shape before averaging, the reported F1 is not the harmonic mean of the reported average precision and recall.

\subsection{Additional Results}
Tab.~\ref{tab:supp_part} extends Tab.~3 with the number of successfully executed shapes and F1. PartPO executes 486 of the 504 programs successfully, compared with 476 for both baselines, and achieves the highest F1. The shape-level reward slightly lowers F1 relative to MeshCoder, because its gain in precision is outweighed by the loss in recall, consistent with the observation in Sec.~\ref{sec:part_eval} that whole-shape optimization alone does not improve the part decomposition.

\begin{table}[t]
\centering
\caption{Extended part-level comparison on the 504 ShapeNetCore test shapes (table, chair and lamp) annotated in PartNet. \#Succ.\ is the number of successfully executed shapes. Part CD is scaled by $10^{3}$; success rate, recall, precision and F1 are in \%. Best results are in bold.}
\label{tab:supp_part}
\resizebox{\columnwidth}{!}{%
\setlength{\tabcolsep}{3pt}
\begin{tabular}{l ccccccc}
\toprule
Method & \#Succ. & Success $\uparrow$ & Part CD $\downarrow$ & Part NC $\uparrow$ & Recall $\uparrow$ & Precision $\uparrow$ & F1 $\uparrow$ \\
\midrule
MeshCoder~\cite{dai2025meshcoder} & 476 & 94.4 & 6.72 & 0.626 & 72.7 & 87.4 & 65.4 \\
\quad + shape-level reward & 476 & 94.4 & 6.10 & 0.627 & 71.8 & 88.4 & 64.9 \\
\quad + \textbf{PartPO} & \textbf{486} & \textbf{96.4} & \textbf{4.74} & \textbf{0.640} & \textbf{74.2} & \textbf{89.2} & \textbf{66.7} \\
\bottomrule
\end{tabular}}
\end{table}

%% ============================================================
%% Supplementary: reconstruction accuracy on commonly executed shapes
%% Requires in the preamble: \usepackage{booktabs}
%% ============================================================

\section{Reconstruction Accuracy on Commonly Executed Shapes}
\label{supp:intersection}

In Tab.~1, CD and ANC are averaged over the shapes that each method executes successfully. Because the methods succeed on different subsets, a method could appear more accurate simply by failing on harder shapes. To rule this out, we recompute both metrics on the intersection of successfully executed shapes, \ie, the test shapes on which all five program-generating methods succeed: 2,505 shapes on ShapeNetCore and 362 on ABO. PrimitiveAnything is not included, since it predicts primitives rather than programs.

Tab.~\ref{tab:supp_intersection} shows that the conclusions hold on this common subset. PartPO achieves the best CD and ANC on both datasets, reducing the CD of MeshCoder from 2.49 to 1.03 on ShapeNetCore and from 2.33 to 1.29 on ABO. It also clearly outperforms the strongest baselines, SPO and the shape-level reward, whose CD remains between 1.73 and 1.93. The improvements of PartPO therefore come from more accurate reconstructions rather than from differences in which shapes are executed successfully.

\begin{table}[t]
\centering
\caption{Reconstruction accuracy on the shapes that all program-generating methods execute successfully (2,505 ShapeNetCore and 362 ABO test shapes). CD is scaled by $10^{2}$. Best results are in bold.}
\label{tab:supp_intersection}
\small
\setlength{\tabcolsep}{4pt}
\begin{tabular}{l cc cc}
\toprule
\multirow{2}{*}{Method} & \multicolumn{2}{c}{ShapeNetCore} & \multicolumn{2}{c}{ABO} \\
\cmidrule(lr){2-3}\cmidrule(lr){4-5}
 & CD $\downarrow$ & ANC $\uparrow$ & CD $\downarrow$ & ANC $\uparrow$ \\
\midrule
MeshCoder~\cite{dai2025meshcoder} & 2.49 & 0.815 & 2.33 & 0.801 \\
\quad + RFT & 2.42 & 0.818 & 2.32 & 0.800 \\
\quad + SPO & 1.84 & 0.825 & 1.73 & 0.817 \\
\quad + shape-level reward & 1.93 & 0.832 & 1.81 & 0.814 \\
\quad + \textbf{PartPO} & \textbf{1.03} & \textbf{0.838} & \textbf{1.29} & \textbf{0.831} \\
\bottomrule
\end{tabular}
\end{table}

%% ============================================================
%% Supplementary: validation of SAM 3 prompt tuning
%% Requires in the preamble: \usepackage{booktabs,multirow}
%% ============================================================

\section{Validation of Prompt Tuning}
\label{supp:prompt_tuning}

The part-semantic reward relies on GT part points obtained from SAM~3 masks (Sec.~3.3). Here we evaluate how much prompt tuning improves these masks over prompting SAM~3 directly with part names.

\paragraph{Setup.}
MeshCoder releases the point clouds of its training data. We run MeshCoder on them and execute the generated programs, keeping only samples whose output mesh is within an L2 Chamfer distance of $10^{-3}$ from the input point cloud, so that the part labels given by the program blocks faithfully describe the shape. We merge fine-grained categories into three classes: \textit{chairs} (armchair, bar chair, chair), \textit{lamps} (lamp, desk lamp) and \textit{tables} (dining table, cocktail table). From these classes, we randomly select 420 samples for training and 120 for testing, and render each sample without texture from six views. Each part of each category has its own set of offset parameters, trained separately on the (part name, mask) pairs of the views in which the part is visible. We use AdamW with a learning rate decayed from $5\times10^{-4}$ to $5\times10^{-6}$ by a cosine schedule, and train each set of parameters for 800 steps. We compare the prompt-tuned model with zero-shot SAM~3 prompted by the part names, using mIoU and mDice on the test views.

\paragraph{Results.}
As shown in Tab.~\ref{tab:supp_prompt_tuning}, prompt tuning improves segmentation substantially on all three classes. The mIoU rises from 18.2\% to 77.2\% on chairs, from 22.6\% to 65.4\% on lamps, and from 59.6\% to 93.6\% on tables. Zero-shot SAM~3 struggles to map fine-grained part names to regions of untextured renders, whereas the tuned prompts align each name with the corresponding geometry. Lamps remain the most difficult class, likely because their parts are often thin or small. Since these test samples come from MeshCoder's synthetic data, the evaluation measures segmentation quality in the training domain of the prompts; Sec.~\ref{supp:grounding} evaluates the resulting 3D part points on real assets.

\begin{table}[t]
\centering
\caption{Part segmentation of SAM~3 with zero-shot part-name prompts and with tuned prompts, on 120 test samples (six views each) rendered from MeshCoder outputs. Values are in \%. \textit{Mean} is the unweighted average over the three classes. Best results are in bold.}
\label{tab:supp_prompt_tuning}
\small
\setlength{\tabcolsep}{5pt}
\begin{tabular}{l cc cc}
\toprule
\multirow{2}{*}{Class} & \multicolumn{2}{c}{mIoU $\uparrow$} & \multicolumn{2}{c}{mDice $\uparrow$} \\
\cmidrule(lr){2-3}\cmidrule(lr){4-5}
 & Zero-shot & Tuned & Zero-shot & Tuned \\
\midrule
Chairs & 18.2 & \textbf{77.2} & 20.4 & \textbf{81.9} \\
Lamps  & 22.6 & \textbf{65.4} & 23.2 & \textbf{69.8} \\
Tables & 59.6 & \textbf{93.6} & 60.6 & \textbf{96.1} \\
\midrule
Mean   & 33.5 & \textbf{78.7} & 34.7 & \textbf{82.6} \\
\bottomrule
\end{tabular}
\end{table}

%% ============================================================
%% Supplementary: quality of GT part grounding
%% Requires in the preamble: \usepackage{amsmath,booktabs}
%% ============================================================

\section{Quality of Part Grounding}
\label{supp:grounding}

The part-semantic reward compares each generated part with GT part points obtained by segmenting rendered views of the GT mesh with the prompt-tuned SAM~3 and lifting the masks to 3D by multi-view voting (Sec.~3.3). Errors in these points directly mislead the reward. Sec.~\ref{supp:prompt_tuning} evaluates the 2D masks on MeshCoder's synthetic outputs; here we evaluate the resulting 3D part points on real assets against human part annotations, and measure how much multi-view voting contributes.

\paragraph{Evaluation data.}
We use the 667 ShapeNetCore training shapes that are also annotated in PartNet~\cite{mo2019partnet}: 264 chairs, 352 tables, 49 lamps and 2 sofas. Each shape is rendered without texture from the same six views as in training, three in each hemisphere. The tuned prompts use a fixed part vocabulary for each category (\eg, \texttt{back}, \texttt{seat} and \texttt{leg} for chairs, and \texttt{table top}, \texttt{leg} and \texttt{stretcher} for tables), whereas PartNet uses its own hierarchical part names. We therefore manually map each vocabulary part to the corresponding PartNet node names, \eg, \texttt{back} to \texttt{chair\_back} and \texttt{table top} to \texttt{tabletop}. For a shape and a vocabulary part $n$, we take the topmost PartNet nodes that match $n$ and use the union of their atomic meshes as the GT geometry of $n$. A vocabulary part without a matching node is not evaluated on that shape, so each shape is evaluated only on the parts that it actually has and that correspond unambiguously. PartNet meshes are aligned with the procedure of Sec.~\ref{supp:part_eval}, using the ShapeNetCore mesh used for rendering as the reference, and we sample PartNet part points $\mathcal Y_n$ densely and uniformly on the GT geometry of each part.

\paragraph{Lifting masks to 3D.}
For each part $n$ and view $v$, SAM~3 predicts a mask with the tuned prompt of $n$. We sample $10^5$ points uniformly on the mesh surface and compare two ways of assigning them to parts. The baseline, \emph{direct back-projection}, assigns a point to part $n$ if it is visible and falls inside the mask of $n$ in at least one view. It uses no cross-view aggregation or filtering, so a segmentation error in any single view, such as a mask that crosses a part boundary, is carried into 3D. \emph{Multi-view voting} is the procedure of Sec.~3.3 that produces the GT part points for training. It ignores observations at grazing angles and accepts a point only if at least $N_{\min}$ views, and at least a fraction $\tau$ of the views that validly observe it, place it inside the mask. We use the same settings as in training, $c_{\min}=0.2$ and $N_{\min}=2$. For a fair comparison, the points of each part are randomly subsampled to at most 2,048 for both methods, the same cap as in training.

\paragraph{Metric.}
Let $\hat{\mathcal X}_n$ be the lifted points of part $n$. We measure the one-sided squared Chamfer distance from $\hat{\mathcal X}_n$ to the PartNet part points,
\begin{equation}
\mathrm{CD}^{\to}(\hat{\mathcal X}_n,\mathcal Y_n)=\frac{1}{|\hat{\mathcal X}_n|}\sum_{x\in\hat{\mathcal X}_n}\min_{y\in\mathcal Y_n}\|x-y\|_2^2,
\end{equation}
in the normalized frame in which the longest side of the bounding box spans $[-1,1]$. This distance measures whether the points assigned to part $n$ actually lie on that part, \ie, the correctness of the GT part points, and the squared distance penalizes points far from the true part more heavily. We use a one-sided distance because the part-semantic reward treats these points as the location of the part: points assigned to the wrong part directly mislead the reward, whereas regions of the true part that are left uncovered, such as surfaces occluded in all views, matter much less.

\paragraph{Results.}
As shown in Tab.~\ref{tab:supp_grounding}, multi-view voting reduces the distance on every category, from 0.574 to 0.302 on tables, from 0.963 to 0.421 on chairs and from 0.874 to 0.311 on lamps, and from 0.750 to 0.351 over all shapes, a reduction of 53\%. Direct back-projection is least accurate on chairs and lamps, whose parts are often adjacent or thin, so that a mask crossing a part boundary in one view mislabels points of a neighboring part; voting removes most of these errors, and the three categories end up at a similar level. The overall distance after voting corresponds to a root-mean-square deviation of about 0.06, or 3\% of the longest side of the bounding box. The GT part points used for training thus agree closely with human annotations on real assets, which complements the evaluation in the synthetic domain in Sec.~\ref{supp:prompt_tuning}.

\begin{table}[t]
\centering
\caption{Quality of the GT part points on the 667 ShapeNetCore training shapes annotated in PartNet. Values are one-sided squared Chamfer distances from the lifted part points to the PartNet parts, scaled by $10^{2}$ (lower is better). \textit{All} is the average over all 667 shapes, including two sofas. Best results are in bold.}
\label{tab:supp_grounding}
\small
\setlength{\tabcolsep}{5pt}
\begin{tabular}{l cccc}
\toprule
Lifting & Table & Chair & Lamp & All \\
\midrule
Direct back-projection & 0.574 & 0.963 & 0.874 & 0.750 \\
Multi-view voting & \textbf{0.302} & \textbf{0.421} & \textbf{0.311} & \textbf{0.351} \\
\bottomrule
\end{tabular}
\end{table}


\end{document}